\documentclass{article}

    \PassOptionsToPackage{numbers,sort,compress}{natbib}

 \usepackage[final]{neurips_2019}

\usepackage[utf8]{inputenc} 
\usepackage[T1]{fontenc}    
\usepackage{hyperref}       
\usepackage{url}            
\usepackage{booktabs}       
\usepackage{amsfonts}       
\usepackage{nicefrac}       
\usepackage{microtype}      

\usepackage{here}    
\usepackage{amsmath}
\usepackage{amssymb}
\usepackage{microtype}
\usepackage{graphicx}
\usepackage{subfigure}
\usepackage{booktabs} 
\usepackage{bm}
\usepackage{amsthm}
\newtheorem{thm}{Theorem}[section]
\newtheorem{lem}[thm]{Lemma}

\newtheorem{as}[thm]{Assumption}
\newtheorem{defi}[thm]{Definition}
\usepackage{color}

\newcommand{\com}[1]{\textcolor{black}{#1}}
\usepackage{wrapfig}

\title{The Normalization Method for Alleviating \\
Pathological Sharpness in Wide Neural Networks}

  \author{Ryo Karakida \\
  AIST \\
  Tokyo, Japan \\
  \texttt{karakida.ryo@aist.go.jp}
   \And
   Shotaro Akaho \\
   AIST \\
  Ibaraki, Japan \\
   \texttt{s.akaho@aist.go.jp} 
   \And
  Shun-ichi Amari \\
  RIKEN CBS \\
  Saitama, Japan\\
  \texttt{amari@brain.riken.jp}
}

\begin{document}

\maketitle

\begin{abstract}
Normalization methods play an important role in enhancing the performance of deep learning while their theoretical understandings have been limited. To theoretically elucidate the effectiveness of normalization, 
we quantify the geometry of the parameter space determined by the Fisher information matrix (FIM), which also corresponds to the local shape of the loss landscape under certain conditions.  
We analyze deep neural networks with random initialization, which is known to suffer from a pathologically sharp shape of the landscape when the network becomes sufficiently wide. We reveal that batch normalization in the last layer contributes to drastically decreasing such pathological sharpness if the width and sample number satisfy a specific condition. 
In contrast, it is hard for batch normalization in the middle hidden layers to  alleviate pathological sharpness in many settings. We also found that layer normalization cannot alleviate pathological sharpness either. 
Thus, we can conclude that batch normalization in the last layer significantly contributes to decreasing the sharpness induced by the FIM.
\end{abstract}

\section{Introduction}
\label{Introduction}
Deep neural networks (DNNs) have performed excellently in various practical applications \cite{nature2015}, but there are still many heuristics and an arbitrariness in their settings and learning algorithms.
To proceed further, it would be beneficial to give theoretical elucidation of how and under what conditions deep learning works well in practice.

Normalization methods are widely used to enhance the trainability and generalization ability of DNNs. In particular, batch normalization makes optimization faster with a large learning rate and achieves better generalization in experiments \cite{ioffe2015batch}. 
Recently, some studies have reported that batch normalization changes the shape of the loss function, which leads to better performance \cite{santurkar2018does,bjorck2018understanding}. 
Batch normalization alleviates a sharp change of the loss function and makes the loss landscape smoother \cite{santurkar2018does}, and prevents an explosion of the loss function and its gradient \cite{bjorck2018understanding}. The  flatness of the loss landscape and its geometric characterization have been explored in various topics such as improvement of generalization ability \cite{keskar2016large,liang2017fisher}, advantage of skip connections \cite{li2018visualizing}, and robustness against adversarial attacks \cite{yao2018hessian}.  Thus, it seems to be an important direction of research to investigate normalization methods from the viewpoint of the geometric characterization. Nevertheless, its theoretical elucidation has been limited to only linear networks \cite{bjorck2018understanding} and simplified models neglecting the hierarchical structure of DNNs \cite{santurkar2018does,lei2016layer}.

One promising approach of analyzing normalization methods is to consider DNNs with random weights and sufficiently wide hidden layers.   
While theoretical analysis of DNNs often becomes intractable because of  hierarchical nonlinear transformations, wide DNNs with random weights can overcome such difficulties and are attracting much attention, especially within the last few years; mean field theory of DNNs \cite{poole2016,schoenholz2016,yang2017,xiao2018dynamical,yang2019bn}, random matrix theory \cite{pennington2018emergence} and kernel methods \cite{daniely2016toward,lee2017deep}. They have succeeded in predicting  hyperparameters with which learning algorithms work well  and even used as a kernel function for the Gaussian process. In addition,  
recent studies on the neural tangent kernel (NTK) have revealed that the Gaussian process with the NTK of random initialization determines even the performance of trained neural networks \cite{jacot2018neural,lee2019wide}. 
Thus, the theory of wide DNNs is becoming a foundation for comprehensive understanding of DNNs.
Regarding the geometric characterization, there have been studies on the Fisher information matrix (FIM) of wide DNNs \cite{karakida2018universal,pennington2018spectrum}. 
The FIM widely appears in the context of deep learning \cite{liang2017fisher,martens2015optimizing}
because it determines the Riemannian geometry of the parameter space and a local shape of the loss landscape around a certain global minimum. In particular, \citet{karakida2018universal} have reported that the eigenvalue spectrum of the FIM is strongly distorted in wide DNNs, that is, the largest eigenvalue takes a pathologically large value (Theorem 2.2). This causes {\it pathological sharpness} of the landscape and such sharpness seems to be harmful from the perspective of optimization \cite{lecun1998efficient} and generalization \cite{keskar2016large}. 

In this study, we focus on the FIM of DNNs and uncover how normalization methods affect it. 
First, to clarify a condition to alleviate the pathologically large eigenvalues, we identify the eigenspace of the largest eigenvalues (Theorem 3.1).
Then, we reveal that batch normalization in the last layer drastically decreases the size of the largest eigenvalues and successfully alleviates the pathological sharpness. This alleviation requires a certain condition on the width and sample size (Theorem 3.3), which is determined by a convergence rate of order parameters.  In contrast, we find that batch normalization in the middle layers cannot alleviate pathological sharpness in many settings (Theorem 3.4) and layer normalization cannot either (Theorem 4.1).
Thus, we can conclude that  batch normalization in the last layer has a vital role in decreasing pathological sharpness. 
Our experiments suggest that such alleviation of the sharpness is helpful in making gradient descent converge even with a larger learning rate.
These results give novel quantitative insight into normalization methods, wide DNNs, and geometric characterization of DNNs and is expected to be helpful in developing a further theory of deep learning.

\section{Preliminaries}
\subsection{Model architecture}

We investigate a fully-connected feedforward neural network with random weights and bias parameters. The network consists of one input layer with $M_0$ units, $L-1$ hidden layers with $M_l$ units per layer ($l=1,2,...,L-1$), and one output layer:
\begin{equation}
u_i^l= \sum_{j=1}^{M_{l-1}} W_{ij}^{l} h_j^{l-1} +b_i^l, \ \ h_i^{l}= \phi(u_i^l), \label{eq1}
\end{equation}
where $h_j^0=x_j$ are inputs. It includes a shallow network ($L=2$) and deep ones ($L \geq 3$). We set the last layer to have a linear readout, i.e., $h_i^{L}=u_i^L$. 
The dimensionality of each variable is given by $W^l \in \mathbb{R}^{M_l\times M_{l-1}}$ and  $h^l, b^l \in \mathbb{R}^{M_l}$.
Suppose that the activation function $\phi(x)$  has a  bounded weak derivative.
A wide class of activation functions, including the sigmoid-like and (leaky-) rectified linear unit (ReLU) functions, satisfy the condition. 
Different layers may have different activation functions. 
Regarding network width, we  set $M_l = \alpha_l M \ \ (l\leq L-1)$ and consider  the limiting case of large $M$ with constant coefficients  $\alpha_l$. The number of readout units is given by a constant $M_L=C$, which is independent of $M$, as is usual in practice. 
Suppose that the parameter set $\theta=\{W_{ij}^{l}, b_i^l \}$ is an ensemble generated by 
\begin{equation}
W_{ij}^l \overset{\text{i.i.d.}}{\sim}\mathcal{N}(0,\sigma_{w}^2/M_{l-1}), \ \  
b_i^l  \overset{\text{i.i.d.}}{\sim}\mathcal{N}(0,\sigma_{b}^2) \label{eq2}
\end{equation}
then fixed, where $\mathcal{N}(0,\sigma^2)$ denotes a Gaussian distribution with zero mean and variance $\sigma^2$.
We assume that there are $T$ input samples $x(t) \in \mathbb{R}^{M_0}$ ($t=1,...,T$) generated from an input distribution  independently and that it is given by a standard normal distribution, i.e.,
\begin{equation}
 x_j(t)   \overset{\text{i.i.d.}}{\sim} \mathcal{N}(0,1). \label{eq7:0410} 
\end{equation}

The FIM  of a DNN is computed by the chain rule in a manner similar to that of the backpropagation algorithm: 
\begin{align}
\frac{\partial f_{k}}{\partial W_{ij}^l} &= \delta_{k,i}^l h_j^{l-1}, \ \   \delta_{k,i}^{l} = \phi'(u_i^l) \sum_{j} \delta_{k,j}^{l+1} W_{ji}^{l+1},
\label{b_chain}
\end{align}
where we denote $f_k=u_k^L$ and $\delta_{k,i}^{l} := \partial f_{k}/\partial u_i^l$ for $k=1,...,C$. To avoid complicated notation, we omit index $k$ of the output unit, i.e.,  $\delta_{i}^{l} = \delta_{k,i}^{l}$.

\subsection{Understanding DNNs through  order parameters}

We use the following four types of {\it order parameters}, i.e., $(\hat{q}^l_t,\hat{q}_{st}^l,\tilde{q}^l_t,\tilde{q}_{st}^l)$, which have been commonly used in various studies of wide DNNs \cite{amari1974method,poole2016,schoenholz2016,yang2017,xiao2018dynamical,lee2017deep,jacot2018neural,lee2019wide,karakida2018universal}.  
First,  we use the following order parameters for feedforward signal propagations: 
$\hat{q}^l_{t} :=  \sum_i h^{l}_i(t)^2 /M_l$ and $\hat{q}^l_{st} := \sum_{i}  h_i^{l} (s) h_{i}^{l}(t) /M_l,$  where $h_i^l(t)$ are the outputs of the $l$-th layer
  when the input is $x_j(t)$ ($t=1,...,T$). 
 The variable $\hat{q}_t^l$ is the total activity of the outputs in the $l$-th layer,  and the 
  variable $\hat{q}^l_{st}$ is the overlap between the activations for different input samples $x(s)$ and $x(t)$. 
 These variables have been utilized to explain the depth to which signals can be sufficiently propagated from the perspective of order-to-chaos phase transition \cite{poole2016}.
 In the large $M$ limit,  these variables are recursively computed by integration over Gaussian distributions \citep{poole2016,amari1974method}: 
\begin{equation}
  \hat{q}^{l+1}_t = \int Du\phi^2 \left( \sqrt{q^{l+1}_t} u \right), \ \ 
\hat{q}^{l+1}_{st} =  I_{\phi}[q^{l+1}_t,q_{st}^{l+1}],    \label{eq_hatq}  
\end{equation}
\begin{equation} 
q^{l+1}_t := \sigma_w^2 \hat{q}^l_t +\sigma_b^2,  \ \ 
q_{st}^{l+1} :=\sigma_w^2 \hat{q}_{st}^l+\sigma_b^2, \label{eq_qst}
\end{equation}
for $l=0, ..., L-1$. Because input samples generated by Eq. (\ref{eq7:0410}) yield 
$\hat{q}^0_t = 1$ and $\hat{q}_{st}^0 = 0$ for all $s$ and $t$,  $\hat{q}_{st}^l$ in each layer takes the same value for all $s \neq t$, and so does $\hat{q}_{t}^l$  for all $t$. The notation $Du = du \exp(-u^2/2) /\sqrt{2\pi}$ means integration over the standard Gaussian density.  
We use a two-dimensional Gaussian integral given by 
$I_{\phi}[a,b] := \int Dy Dx \phi (\sqrt{a}x)\phi (\sqrt{a}(cx +\sqrt{1-c^2}y ))$ with $c=b/a$. 

We also use  the corresponding variables for backward signals:
    $ \tilde{q}^l_{t} := \sum_i \delta_i^l(t)^2$ and $\tilde{q}^l_{st} := \sum_{i}  \delta_i^{l} (s) \delta_{i}^{l}(t). $ The variable 
 $\tilde{q}^l_t$ is the magnitude of the backward signals and 
 $\tilde{q}^l_{st}$ is their overlap. Previous studies found that $\tilde{q}^l_{st}$ and $\tilde{q}^{l}_{st}$ in the large $M$ limit are easily computed using the following recurrence relations \cite{schoenholz2016,yang2019scaling},  
\begin{align} 
\tilde{q}^l_t = \sigma_w^2 \tilde{q}^{l+1}_t \int Du \left[\phi' (\sqrt{q^{l}_t}u) \right]^2,   \ \
\tilde{q}_{st}^l =  \sigma_w^2 \tilde{q}_{st}^{l+1}  I_{\phi'}[q^l_t,q_{st}^l],  \label{eq_tilqst} 
\end{align}
 for $l=0,...,L-1$ with $\tilde{q}^L_t =\tilde{q}_{st}^L=1$.
 Previous studies confirmed excellent agreements between these backward order parameters and experimental results \cite{schoenholz2016,yang2017,xiao2018dynamical}. Although these studies required the so-called {\it gradient independence assumption} to derive these recurrences (details are given in Assumption 3.2), \citet{yang2019scaling} has recently proved that such assumption is unnecessary when $\phi(x)$ has a polynomially bounded weak derivative.

The order parameters depend only on $\sigma_w^2$ and $\sigma_b^2$, the types of activation functions, and depth. 
The recurrence relations require $L$ iterations of one- and two-dimensional numerical integrals. They are analytically tractable in certain activation functions including the ReLUs \cite{karakida2018universal}. 




\subsection{Pathological sharpness of local landscapes}

 The FIM  plays an essential role in the geometry of the parameter space and is a fundamental quantity in both statistics and machine learning. It defines a Riemannian metric of the parameter space, where the infinitesimal difference of statistical models is measured by Kullback-Leibler divergence, as in information geometry \cite{amari2016}. 
 We analyze the eigenvalue statistics of the following FIM of DNNs \cite{park2000,pascanu2013,pennington2018spectrum,karakida2018universal},
 \begin{equation}
F =  \sum_{k=1}^C \mathrm{E}[ \nabla_\theta f_{k}(t) \nabla_\theta f_k (t)^\top], \label{eqFIM}
\end{equation}
where $\theta$ is a vector composed of all parameters $\{W_{ij}^l,b_i^l\}$ and  $\nabla_\theta$ is the derivative with respect to it. 
The average over an input distribution is denoted by $\mathrm{E}[\cdot]$. As usual, 
when $T$ input samples $x(t)$ $(t=1,...,T)$ are available for training, we replace the expectation $\mathrm{E}[\cdot]$ with the FIM by the empirical average over $T$ samples, i.e., $\mathrm{E}[\cdot] = \frac{1}{T}\sum_{t=1}^T$.
This study investigates such an empirical FIM for arbitrary $T$.
It converges to the expected FIM as $T \rightarrow \infty$. 
This empirical FIM is widely used in machine learning and corresponds to the statistical model for the squared-error loss \cite{park2000,pascanu2013,pennington2018spectrum} (see \citet{karakida2018universal} for more details on this FIM).
\com{Recently,  \citet{kunstner2019limitations} emphasized that in the context of natural gradient algorithms,  the FIM (\ref{eqFIM}) leads to better optimization than an FIM approximated by using training labels.}

The FIM is known to determine not only the local distortion of the parameter space but also the loss landscape around a certain global minimum. 
Suppose the squared loss function $E(\theta)= (1/2T)\sum_{k=1}^C \sum_{t=1}^T(y_k(t)-f_k(t))^2$, where $y_k(t)$ represents a training label corresponding to the input sample $x(t)$.  The FIM is related to the Hessian of the loss function, $H := \nabla_{\theta} \nabla_{\theta} E(\theta)$,  in the  following manner \cite{karakida2018universal,pennington2018spectrum}:
\begin{equation}
H = F - \frac{1}{T} \sum_t^T \sum_k^C (y_k(t)-f_{k}(t)) \nabla_{\theta} \nabla_{\theta} f_{k}(t). \label{eq_9}
\end{equation}
The Hessian coincides with the empirical FIM when the parameter converges to the global minimum with zero training error.
\com{In that sense, the FIM determines the local shape of the loss landscape around the minimum.} 
This FIM is also known as the Gauss-Newton approximation of the Hessian. 

\citet{karakida2018universal}  elucidated hidden relations between the order parameters and basic statistics of the FIM's eigenvalues. We investigate DNNs satisfying the following condition.
\begin{defi}
{\it Suppose a DNN with bias terms ($\sigma_b \neq 0$) or activation functions satisfying the non-zero Gaussian mean. We refer to this as a non-centered network.  
}
\end{defi}
The definition of the non-zero Gaussian mean is $\int Dz \phi(z) \neq 0$.  Non-centered networks include various realistic settings because usual networks include bias terms, and widely used activation functions, such as the sigmoid function and (leaky-) ReLUs, have the non-zero Gaussian mean. 
Denote the FIM's eigenvalues  as $\lambda_i$ ($i=1,...,P$) where $P$ is the number of all parameters. The eigenvalues are non-negative by definition. Their mean is $m_{\lambda}:=\sum_i^P\lambda_i/P$ and the maximum is $\lambda_{max}:=\max_i \lambda_i$. The following theorem holds:
\begin{thm}[\cite{karakida2018universal}]
{\it Suppose a non-centered network and i.i.d. input samples generated by Eq.  (\ref{eq7:0410}).
When $M$ is sufficiently large, the eigenvalue statistics of $F$ are asymptotically evaluated as    
\begin{align}
m_\lambda  &\sim  \kappa_1 C/M,  \ \
\lambda_{max}\sim \alpha \left( \frac{T-1}{T} \kappa_2 + \frac{\kappa_1 }{T}\right)M,
\end{align}
where $\alpha := \sum_{l=1}^{L-1} \alpha_l \alpha_{l-1}$, and positive constants $\kappa_1$ and $\kappa_2$ are obtained using order parameters, 
\begin{equation}
\kappa_1 :=  \sum_{l=1}^L\ \frac{\alpha_{l-1}}{\alpha} \tilde{q}^l_t  \hat{q}^{l-1}_t, \ \ \kappa_2 := \sum_{l=1}^L   \frac{\alpha_{l-1}}{\alpha} \tilde{q}^l_{st}\hat{q}^{l-1}_{st}.
\end{equation}
}
\end{thm}
The mean is asymptotically close to zero and it implies that most of the eigenvalues are very small. In contrast, $\lambda_{max}$ becomes pathologically large in proportion to the width. We refer to this $\lambda_{max}$  as {\it pathological sharpness} since FIM's eigenvalues determine the local shape of the parameter space and loss landscape.
Empirical experiments reported that both of close-to-zero eigenvalues and pathologically large ones appear in trained networks as well \cite{lecun1998efficient,sagun2017empirical}. 

Pathological sharpness universally appears in various DNNs. Technically speaking, if the network is {\it not} non-centered (i.e., a network with no bias terms and zero-Gaussian mean; we call it a {\it centered network}),  $\kappa_2=0$ holds and lower order terms of the eigenvalue statistics become non-negligible \cite{karakida2018universal}, and the pathological sharpness may disappear. For instance, $\lambda_{max}$ is of $O(1)$ when $T$ is properly scaled with $M$ in a centered shallow network   \cite{pennington2018spectrum}.
Except for such special centered networks, we cannot avoid pathological sharpness. 
In practice, it would be better to alleviate  the pathologically large $\lambda_{max}$ because it causes the sharp loss landscape. It requires very small learning rates (see Section 3.4) and will lead to worse generalization \cite{keskar2016large,bjorck2018understanding}. 
In the following section, we reveal that a specific normalization method plays an important role in alleviating pathological sharpness.

\section{Alleviation of pathological sharpness in batch normalization}

\subsection{Eigenspace of largest eigenvalues}

Before analyzing the effects of normalization methods on the FIM, it will be helpful to characterize the cause of pathological sharpness. We find 
the following eigenspace of $\lambda_{max}$'s:
\begin{thm}
{\it Suppose a non-centered network and i.i.d. input samples generated by Eq.  (\ref{eq7:0410}).
When $M$ is sufficiently large, the eigenvectors corresponding to $\lambda_{max}$'s are asymptotically equivalent to  
\begin{equation}
    \mathrm{E}[\nabla_\theta f_k] \ \ (k=1,...,C).
\end{equation}
}
\end{thm}
The derivation is shown in Supplementary Material A.2. 
This theorem gives us an idea of the effect of normalization on the FIM. Assume that we could shift the model output as $\bar{f}_k(t) = f_k(t) -  \mathrm{E}[f_k]$. In this shifted model, the eigenvectors become $ \mathrm{E}[\nabla_\theta \bar{f}_k]=0$ and vanish. Naively thinking,  pathologically large eigenvalues may disappear under this shift of outputs. The following analysis shows that this naive speculation is correct in a certain condition.

\subsection{Batch normalization in last layer}

In this section, we analyze batch normalization in the last layer ($L$-th layer): 
\begin{align}
f_k(t) &:= \frac{u_k^L (t) - \mu_k(\theta)}{\sigma_k(\theta)}\gamma_k+\beta_k, \label{eq13:0513} \\
 \mu_k(\theta)  &:= \mathrm{E}[u_k^L (t)], \ \
 \sigma_k(\theta) := \sqrt{\mathrm{E}[u_k^L (t)^2]-\mu_k(\theta)^2}, 
\end{align}
for $k=1,...,C$. The average operator $\mathrm{E}[\cdot]$ is taken over all input samples. 
 In practical use of batch normalization in stochastic gradient descent, the training samples are often divided into many small mini-batches, but we do not consider such division since our current interest is to evaluate the FIM averaged over all samples. 
We set  the hyperparameter  $\gamma_k=1$  for simplicity because $\gamma_k$ only changes the scale of the FIM up to a constant. The  constant $\beta_k$  works as a new bias term in the normalized network.
We do not normalize middle layers ($1 \leq l \leq L-1$) to observe only the effect of normalization in the last layer. 

In the following analysis, we use a widely used assumption for DNNs with random weights:
\begin{as}[the gradient independence assumption \cite{schoenholz2016,yang2017,xiao2018dynamical,yang2019bn,karakida2018universal}]
{\it When one evaluates backward order parameters,  one can replace 
weight matrices $W^{l+1}$ in the chain rule (\ref{b_chain}) with a fresh i.i.d. copy, i.e.,  $\tilde{W}^{l+1}_{ij}  \overset{\text{i.i.d.}}{\sim}\mathcal{N}(0,\sigma_{w}^2/M_{l})$.}
\end{as}
Supposing this assumption has been a central technique of the mean field theory of DNNs  \cite{schoenholz2016,yang2017,xiao2018dynamical,karakida2018universal,yang2019bn} to make the derivation of backward order parameters relatively easy. 
These studies confirmed that this assumption leads to excellent agreements with experimental results. Moreover, recent studies \cite{yang2019scaling,arora2019exact} have succeeded in theoretically justifying that various statistical quantities obtained under this assumption coincide with exact solutions. Thus, Assumption 3.2 is considered to be effective as the first step of the analysis.

First, let us set $\sigma_k(\theta)$ as a constant and only consider mean subtraction in the last layer:
\begin{align}
\bar{f}_k(t) &:= (u_k^L (t) - \mu_k(\theta))\gamma_k+\beta_k.   \label{outputNorm} 
\end{align}
Since the $\sigma_k(\theta)$ controls the scale of the network output, one may suspect that the contribution of the mean subtraction would only be restrictive for alleviating  sharpness. Contrary to this expectation, we find an interesting fact that the mean subtraction is essential to alleviate pathological sharpness: 
\begin{thm}
{\it  Suppose a non-centered network with the mean subtraction in the last layer (Eq. (\ref{outputNorm})) and i.i.d. input samples generated by Eq. (\ref{eq7:0410}).  In the large $M$ limit,  the mean of the FIM's eigenvalues  is asymptotically evaluated by
\begin{equation}
    m_\lambda \sim (1-1/T)(\kappa_1-\kappa_2)  C/M. \label{eq16:0520}
\end{equation}
The largest eigenvalue is asymptotically evaluated as follows: (i) when $T \geq 2$ and $T=O(1)$,
\begin{equation}
    \lambda_{max}  \sim \alpha  \frac{\kappa_1-\kappa_2}{T}M, \label{eq17:0512}
\end{equation}
and (ii) when $T=O(M)$ with a constant  $\rho:=M/T$, under the gradient independence assumption, we have
\begin{equation}
   \rho \alpha (\kappa_1 - \kappa_2) +  c_1   \leq  \lambda_{max}  \leq \sqrt{ (C\alpha^2 \rho (\kappa_1 - \kappa_2)^2 + c_2) M}, \label{eq19:0512}
\end{equation}
for non-negative constants $ c_1$ and $c_2$.
}
\end{thm}
The derivation is shown in Supplementary Material B.1. The mean subtraction does not change the order of $\lambda_{max}$ when $T=O(1)$.  In contrast, it is interesting that it decreases the order when $T=O(M)$.
The decrease in $m_\lambda$ only appears in the coefficient because  $\kappa_1>\kappa_1-\kappa_2>0$ hold in the non-centered networks.  Thus, we can conclude that the mean subtraction in the last layer plays an essential role in decreasing $\lambda_{max}$ when $T$ is appropriately scaled to $M$.

As shown in Fig.1, we empirically confirmed that $\lambda_{max}$ became of $O(1)$ in numerical experiments and pathological sharpness disappeared when $T=M$. Theorem 3.3 is consistent with the numerical experimental results. 
We numerically computed $\lambda_{max}$ in DNNs with random Gaussian weights, biases, and input samples generated by Eq. (\ref{eq7:0410}). We set $\alpha_l=C=1$ and $L=3$. Variances of parameters were given by ($\sigma_w^2,\sigma_b^2$) = ($2,0$) in the ReLU case, and ($3,0.64$) in the tanh case.  Each points and error bars show the experimental results over $100$ different ensembles.
 We show the value of $\rho \alpha (\kappa_1-\kappa_2)$ as the lower bound of $\lambda_{max}$ (red line). 
Although this lower bound and the theoretical upper bound of order  $\sqrt{M}$ are relatively loose compared to the experimental results, recall that our purpose is not to obtain the tight bounds but to show the alleviation  of $\lambda_{max}$. The experimental results with the mean subtraction were much lower than those without it as our theory predicts.

From a theoretical perspective, one may be interested in how Assumption 3.2 works in the evaluation of $\lambda_{max}$.  
As shown in Theorem B.1 of Supplementary Material B.1,
we can evaluate $\lambda_{max}$ even without using this assumption. In general, the mean subtraction in the last layer makes $\lambda_{max}$ depend on a convergence rate of backward order parameters. That is, when   
we have $\sum_{i}  \delta_i^{l} (s) \delta_{i}^{l}(t)= \tilde{q}^l_{st}+O(1/M^q)$ with the convergence rate $q>0$, it leads to $O(M^{1-2q^*}) \leq \lambda_{max} \leq O(M^{1-q^*})$ with $q^*=\min \{1/2,q \}$. This means that the alleviation appears for any $q$.
In particular, Assumption 3.2 yields $q=q^*=1/2$ and we obtain the lower bound of order $1$.  We have also confirmed that backward order parameters in numerical experiments on DNNs with random weights achieved the convergence rate of $q=q^*=1/2$ (in Fig. S.1).  
Thus, Theorem 3.3 under this assumption becomes consistent with the experimental results.
\com{The batch normalization essentially requires the evaluation of the convergence rate and this is an important difference from the previous study on DNNs without normalization methods \cite{karakida2018universal}.}

We can also add $\sigma_k^L(\theta)$ to Theorem 3.3 and obtain the eigenvalue statistics under the normalization (Eq. (\ref{eq13:0513})). When $T=O(M)$, the eigenvalue statistics slightly change to 
\begin{align}
    m_\lambda &\sim Q_1 (\kappa_1-\kappa_2)/M, \ \
   \rho \alpha \frac{Q_2}{Q_1} (\kappa_1 - \kappa_2) +  c_1'  \leq  \lambda_{max}  \leq \sqrt{ (Q_2 \alpha^2 \rho (\kappa_1 - \kappa_2)^2 + c_2') M} \label{eq19:190523}
\end{align}
where $Q_1:= \sum_k^C 1/\sigma_k(\theta)^2$, $Q_2:= \sum_k^C 1/\sigma_k(\theta)^4$,  $c_1'$ and $c_2'$ are non-negative constants. 
The derivation is shown in Supplementary Material B.2. 
This clarifies that the variance normalization works only as a constant factor and the mean subtraction is essential to reduce pathological sharpness.



\begin{figure}
\vskip 0.2in
\begin{center}
\centerline{\includegraphics[width=0.80\columnwidth]{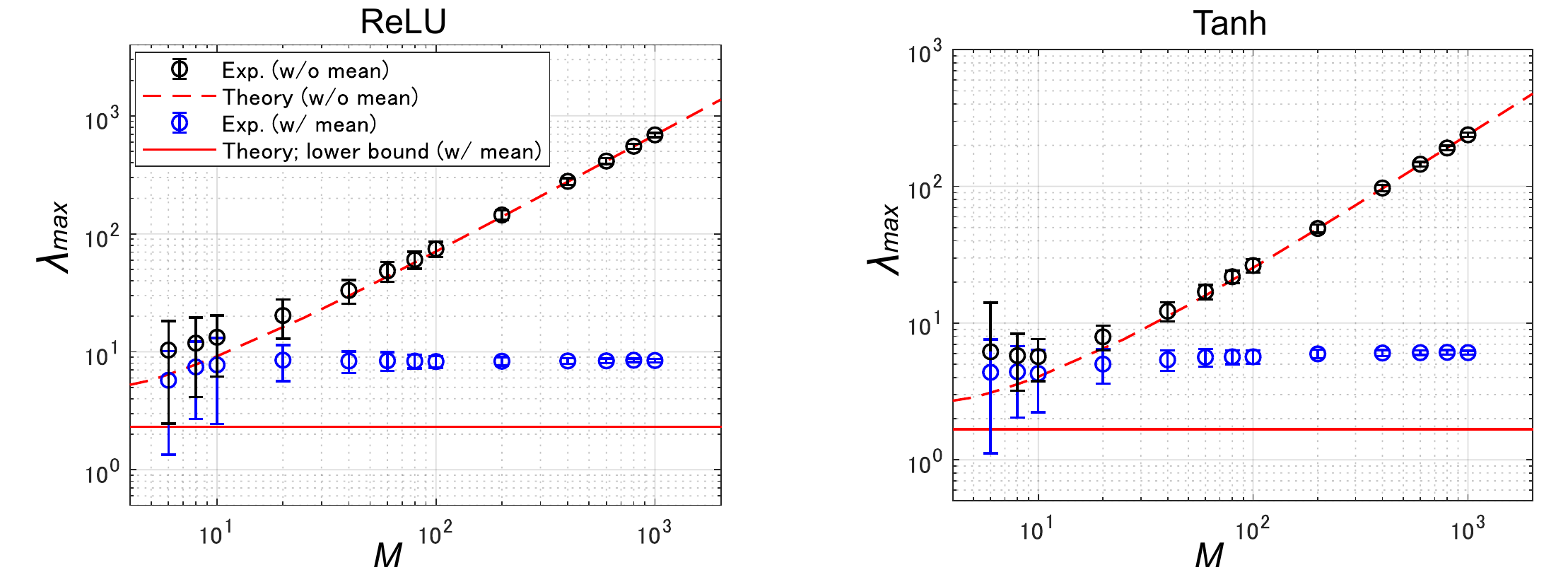}}
\caption{Effect of mean subtraction in last layer on $\lambda_{max}$.  Largest eigenvalues with $T=M$ are shown. Black points show experimental values without mean subtraction, and red dashed lines show theoretical values of $\lambda_{max}$ in Theorem 2.2. In contrast, blue points show experimental values with mean subtraction and red solid lines show theoretical values of lower bound in Theorem 3.3.  
}
\label{fig1}
\end{center}
\vskip -0.2in
\end{figure}

\subsection{Batch normalization in middle layers}

To distinguish the effectiveness of normalization in the last layer from those in other layers,
we apply batch normalization in all layers {\it except for} the last layer :
\begin{align}
u_i^l(t)&= \sum_{j=1}^{M_{l-1}} W_{ij}^{l} h_j^{l-1}(t) +b_i^l, \ \ 
\bar{u}^l_i(t) := \frac{u_i^l (t) - \mu_i^l}{\sigma_i^l} \gamma^l_i + \beta_i^l,  \ \  h_i^l(t) = \phi(\bar{u}^l_i(t)),  \label{Norm} \\ 
\mu_i^l&:= \mathrm{E}[ u_i^l (t)], \ \   
\sigma_i^l := \sqrt{\mathrm{E}[ u_i^l (t)^2] - (\mu_i^l)^2}, 
\end{align}
for all middle layers $(l=1,...,L-1)$ while the last layer is kept in an un-normalized manner, i.e.,  $f_k(t) = \sum_{j} W_{kj}^{L} h_j^{L-1}(t) +b_k^L$. The variables $\mu_i^l$ and $\sigma_i^l$ depend on weight and bias parameters. 
For simplicity, we set $\gamma_i^l=1$ and $\beta_i^l=0$. 
We find a lower bound of $\lambda_{max}$ with order of $M$:
\begin{thm}\label{colN1}
{\it Suppose non-negative activation functions and i.i.d. input samples generated by Eq. (\ref{eq7:0410}).  
 The largest eigenvalue of the FIM under the normalization (Eq. (\ref{Norm})) is asymptotically lower bounded by
\begin{equation}
\lambda_{max} \geq  \alpha_{L-1} \left( \frac{T-1}{T}\hat{q}^{L-1}_{st,BN}+ \frac{\hat{q}^{L-1}_{t,BN}}{T}  \right) M,
\end{equation}
where $\hat{q}^{L-1}_{t,BN}$ and $\hat{q}^{L-1}_{st,BN}$ are positive constants independent of $M$.
}
\end{thm}
Because the last layer is unnormalized, we can construct a lower bound composed of the  activations in the $(L-1)$-th layer. Note that the set of non-negative activation functions (i.e., $\phi(x)\geq 0$) is a subclass of the non-centered networks. It includes sigmoid and ReLU functions which are widely used. 
The bias term, i.e., $\sigma_b^2$, does not affect the theorem because they are canceled out in the mean subtraction of each middle layer. 
 After this batch normalization,  $\lambda_{max}$ is still of $O(M)$ at lowest and the pathological sharpness is unavoidable in that sense. Thus, one can conclude that the normalization in the middle layers cannot alleviate pathological sharpness in many settings. 

The constants  $\hat{q}^{L-1}_{t,BN}$ and $\hat{q}^{L-1}_{st,BN}$ correspond to feedforward order parameters in batch normalization. The details are shown in  Supplementary Material C.1.  
Although the purpose of our study was to evaluate the order of the eigenvalues,  
 some approaches analytically compute the specific values of the order parameters under certain conditions \cite{yang2019bn,wei2019meanfield} (see Supplementary Material C.2 for more details).  
In particular, they are analytically tractable in ReLU networks as follows; $ \hat{q}^{L-1}_{t,BN}=1/2$  and  $\hat{q}^{L-1}_{st,BN} = \frac{1}{2} J(-1/(T-1))$ where $J(x)$ is the arccosine kernel \cite{yang2019bn}. 

\subsection{Effect on the gradient descent method}

Consider the gradient descent method in a batch regime.  Its update rule is given by 
$\theta_{t+1} \leftarrow \theta_t  -\eta \nabla_\theta E(\theta_t)$  where $\eta$ is a constant learning rate. 
 Under some natural assumptions, there exists a necessary condition of the learning rate for the gradient dynamics to converge to a global minimum \cite{lecun1998efficient,karakida2018universal}; 
 \vspace{-3pt}
\begin{equation}
\eta <  2/\lambda_{max}. \label{eq23:1016}
\end{equation}
Because our theory shows that batch normalization in the last layer decreased $\lambda_{max}$, the appropriate learning rate for convergence becomes larger. To confirm this effect on the learning rate, we did experiments on training with the gradient descent as shown in Fig. 2. we trained DNNs with various widths by using various fixed learning rates, providing i.i.d. Gaussian input samples and labels generated by corresponding teacher networks. It was the same setting as the experiment shown in \cite{karakida2018universal}. Fig. 2 (left) shows the color map of training losses without any normalization method and is just a reproduction of  \cite{karakida2018universal}.  Losses exploded in the gray area (i.e., were larger than $10^3$) and the red line shows the theoretical value of $2/\lambda_{max}$, which was calculated with the FIM at random initialization. Training above the red line exploded in sufficiently widen DNNs, just as the necessary condition (\ref{eq23:1016}) predicts. In contrast, Fig. 2 (right) shows the result of the batch normalization (mean subtraction) in the last layer. We confirmed that it allows larger learning rates for convergence and they are independent of width.
We calculated the theoretical line by using the lower bound of $\lambda_{max}$, i.e., $\eta = 2/(\rho \alpha (\kappa_1-\kappa_2))$. Note that Fig. 2 shows the results on the single trial of training with  fixed initialization. It caused the stripe pattern of color map depending on the random seed of each width, especially in the case of normalized networks. 
As shown in Fig. S.2 of Supplementary Material D,  accumulation of multiple trials achieves lower losses regardless of the width.  
Thus, the batch normalization is helpful to set larger learning rates, which could be expected to speed-up the training of neural networks \cite{ioffe2015batch}.

\begin{figure}
\vspace{-5pt}
\centering
\includegraphics[width=0.65\textwidth]{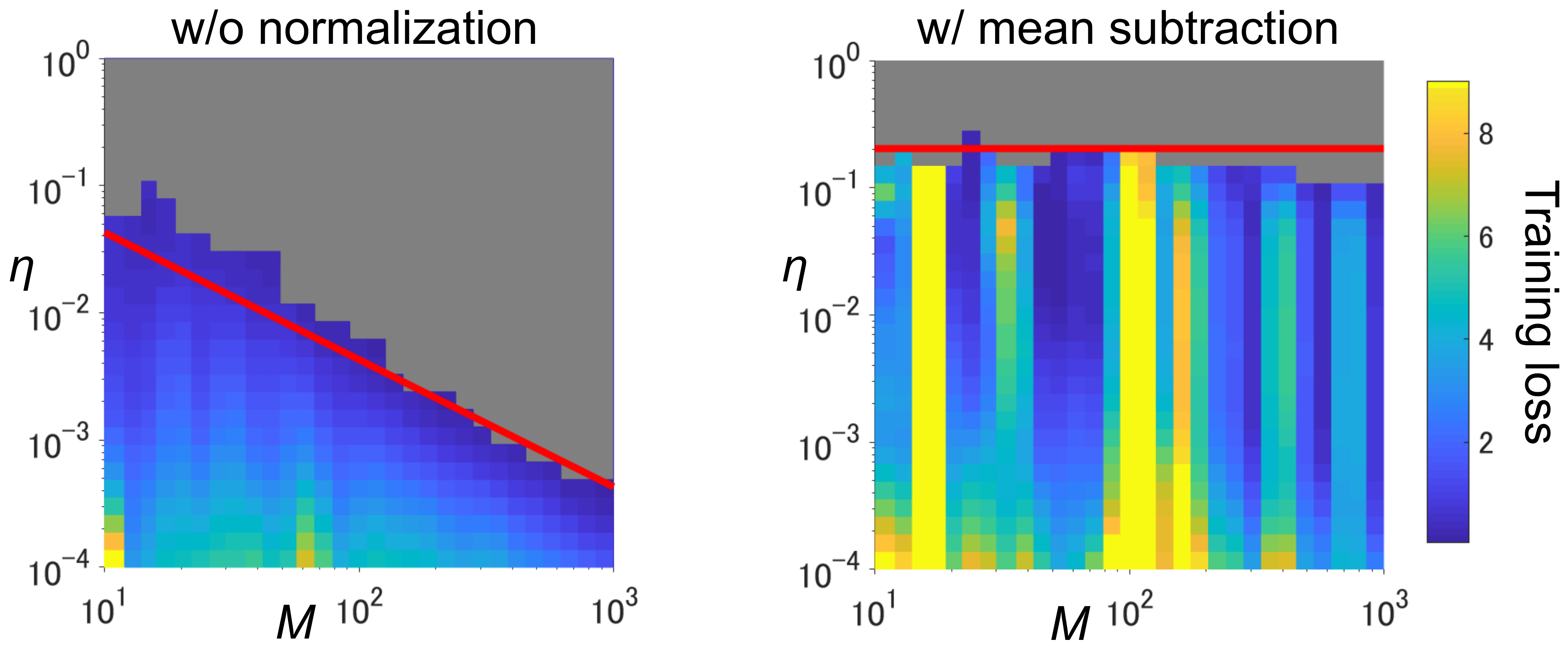}
\caption{Exhaustively searched training losses depending on $M$ (width) and $\eta$ (learning rate). The color bar shows the value of training loss after $1000$ steps of training. We trained deep ReLU networks with $\alpha_l=C=1$, $L=3$, $T=1000$ and ($\sigma_w^2,\sigma_b^2)=(4,1)$. }
\vspace{-5pt}
\end{figure}

\section{Pathological sharpness in layer normalization}
It is an interesting question to investigate the effect of other normalization methods on pathological sharpness.
Let us consider layer normalization \cite{lei2016layer}:
\begin{align}
u_i^l(t)&= \sum_{j=1}^{M_{l-1}} W_{ij}^{l} h_j^{l-1}(t) +b_i^l, \ \ 
\bar{u}^l_i(t) := \frac{u_i^l (t) - \mu^l(t)}{\sigma^l(t)} \gamma^l_i + \beta_i^l,  \ \  h_i^l = \phi(\bar{u}^l_i(t)), \label{LNorm} \\ 
\mu^l(t)&:= \sum_i u_i^l (t)/M_l, \ \   
\sigma^l(t) := \sqrt{ \sum_i u_i^l (t)^2/M_l -\mu^l(t)^2},
\end{align}
for all layers ($l=1,...,L$). The network output is normalized as  $f_k(t)=\bar{u}^L_k(t)$.
\com{While batch normalization (\ref{Norm}) normalizes the pre-activation of each unit across batch samples, layer normalization (\ref{LNorm}) normalizes that of each sample across the units in the same layer.}
Although layer normalization is the method typically used in recurrent neural networks, we show its effectiveness in feedforward networks to contrast the effect of batch normalization on the FIM.  
For simplicity, we set $\gamma_i^l=1$ and $\beta_i^l=0$. Then, we find 
\begin{thm}
{\it  Suppose a non-centered network, i.i.d. input samples generated by Eq. (\ref{eq7:0410}), and the gradient independence assumption.
When $M$ is sufficiently large and $C>2$, the eigenvalue statistics of the FIM under the normalization (Eq. (\ref{LNorm})) are asymptotically evaluated as    
\begin{align}
m_\lambda  &\sim (C-2) \eta_1 \kappa_1' /M,  \ \
\frac{ \alpha s}{(C-2)\eta_1\kappa_1' }  M \leq \lambda_{max} \leq  \alpha \sqrt{s}M, 
\end{align}
with
\begin{equation}
    s= \frac{(\eta_3^2-\eta_1^2)T+(C-2)(\eta_1^2T-\eta_2)}{T}{\kappa'_2}^{2}+(C-2)\eta_2\frac{{\kappa'_1}^{2}}{T},
\end{equation}
where $\kappa_1'$, $\kappa_2'$ and $\eta_i$ $(i=1,2,3)$ are constants independent of $M$.
}
\end{thm}
Layer normalization does not alleviate the pathological sharpness in the sense that $\lambda_{max}$ is of order $M$. Intuitively, this is because $\mathrm{E}[f_k]$ is not equal to $\sum_k f_k/C$ and the mean subtraction in the last layer does not cancel out the eigenvectors in Theorem 3.1. 
 We can compute $\kappa_1'$ and $\kappa_1'$ by using order parameters and $\eta_i$ $(i=1,2,3)$ by the variance $\sigma^L(t)^2$. The definition of each variable and the proof of the theorem  are shown in Supplementary Material E.
 The independence assumption is used in derivation of backward order parameters as usual \cite{schoenholz2016,yang2019bn}. 
When $C=2$, the FIM becomes a zero matrix because of a special symmetry in the last layer. Therefore, the non-trivial case is $C>2$.


\section{Related work}

\vspace{-3pt}

{\bf Normalization and  geometric characterization.}
Batch normalization is believed to perform well  because it suppresses the internal covariate shift \cite{ioffe2015batch}. Recent extensive studies, however, have reported alternative explanations on how  batch normalization works \cite{santurkar2018does,bjorck2018understanding}. 
\citet{santurkar2018does} empirically found that batch normalization decreases a sharp change of the loss function and makes the loss landscape smoother. \citet{bjorck2018understanding} reported that batch normalization works to prevent an  explosion of the loss and gradients.
While some theoretical studies analyzed FIMs in un-normalized DNNs \cite{pennington2018spectrum,karakida2018universal,liang2017fisher}, analysis in normalized DNNs has been limited.  
\citet{santurkar2018does} analyzed gradients and Hessian under batch normalization in a single layer and theoretically evaluated their worst case bounds, but its inequality was too general to quantify the decrease of sharpness. In particular, it misses the special effect of the last layer, as we found in this study.
The original paper \cite{lei2016layer} of layer normalization analyzed the FIM in generalized linear models (GLMs) and argued that the normalization could decrease curvature of the parameter space. While a GLM corresponds to the single layer model,  shallow and deep networks have hidden layers. As the hidden layers become wide,  pathological sharpness appears and layer normalization suffers from it. 

{\bf Gradient descent method.}  
There are other related works in addition to those mentioned in Section 3.4. \citet{bjorck2018understanding} speculated that larger learning rates realized by batch normalization may help stochastic gradient descent avoid sharp minima and it leads to better generalization.  
\citet{wei2019meanfield} estimated $\lambda_{max}$ and $\eta$ under a special type of batch-wise normalization. Because their normalization method approximates a chain rule of backpropagation by neglecting the contribution of mean subtraction, it suffers from pathological sharpness and requires smaller learning rates.

{\bf Neural tangent kernel.} The FIM and NTK satisfy a kind of duality, and share  the same non-zero eigenvalues. Our proofs on the eigenvalue statistics use NTK with standard parameterization, i.e., $F^*$ in Supplementary Material A.1. The NTK at random initialization is known to determine the gradient dynamics of a  sufficiently wide DNN in function space. The sufficiently wide network can achieve a zero training error and it means that there is always a global minimum sufficiently close to random initialization.
In the parameter space, \citet{lee2019wide} proved that NTK dynamics is sufficiently approximated by the gradient descent of a linearized model expanded around  random initialization $\theta_0$:
$f(x;\theta_t) = f(x;\theta_0) + \nabla_\theta f (x;\theta_0)^\top \omega_t$,
where $\omega_t := \theta_t-\theta_0$ and $t$ means the step of the gradient descent. 
Naively speaking, this suggests that the optimization of the wide DNN approximately becomes convex and the loss landscape is dominated by a  quadratic form with the FIM, i.e., $ \omega_t^\top F \omega_t$.


\section{Discussion}

There remain a number of directions for extending our theoretical framework. Recent studies on wide DNNs have revealed that the NTK of random initialization dominates  the training dynamics and even the performance of trained networks \cite{jacot2018neural,lee2019wide}. 
Since the NTK is defined as a right-to-left reversed Gram matrix of the FIM under a  special parameterization, the convergence speed of the training dynamics is essentially governed by the eigenvalues of the FIM at the random initialization. Analyzing these dynamics under normalization remains to be uncovered. For further analysis, random matrix theory will also be helpful in obtaining the whole eigenvalue spectrum or deriving tighter bounds of the largest eigenvalues. Although random matrix theory has been limited to a single layer or shallow networks \cite{pennington2018spectrum}, it will be an important direction to extend it to deeper and normalized networks.

There may be potential properties of normalization methods that are not detected in our framework. 
\citet{kohler2018towards} analyzed  the decoupling of the weight vector to its direction and length as in batch normalization and weight normalization. They revealed that such decoupling could contribute to accelerating the optimization.  
 \citet{bjorck2018understanding} discussed that deep linear networks without bias terms suffer from the explosion of the feature vectors and speculated that batch normalization is helpful in reducing this explosion. This implies that batch normalization may be helpful to improve optimization performance even in a centered network. 
\citet{yang2019bn}  developed an excellent mean-field framework for batch normalization through all layers and found that the gradient explosion is induced by batch normalization in networks with extreme depth. Even if batch normalization alleviates pathological sharpness regarding the width, the coefficients of order evaluation can become very large when the network is extremely deep. It may cause another type of sharpness.  
\com{
It is also interesting to explore SGD training under normalization and quantify how the alleviation of sharpness affects appropriate sizes of learning rate and mini-batch, which have been mainly investigated in SGD training without normalization \cite{park2019effect}. 
}
Further studies on such phenomena in wide DNNs would be helpful for further understanding and development of normalization methods.

\newpage

\subsubsection*{Acknowledgments}
This work was partially supported by a Grant-in-Aid for Young Scientists (19K20366) from the Japan Society for the Promotion of Science (JSPS).

\bibliographystyle{unsrtnat}
\bibliography{test}

\newpage

\setcounter{section}{0}
\renewcommand{\thesection}{\Alph{section}}
\renewcommand{\theequation}{S.\arabic{equation}}
\setcounter{equation}{0}

\part*{Supplementary Materials}

\section{Basic statistics of FIM without normalization}

\subsection{Reversed FIM}
We prepare the following two lemmas to prove the theorems in the main text.

An FIM is a $P \times P$ matrix, where $P$ is the dimension of all parameters. 
Define a $P\times CT$ matrix $R$ by  
\begin{eqnarray}
R &:=& 
\frac{1}{\sqrt{T}}[\nabla_\theta f_1 \ \ \nabla_\theta f_2 \ \  \cdots \ \  \nabla_\theta f_C]. \label{eq60:0111}
\end{eqnarray}
Its columns are the gradients on each input, i.e., $\nabla_{\theta} f_k(t)$ $(t=1,...,T)$. 
One can represent an empirical FIM by 
\begin{equation}
F=RR^{\top}.    
\end{equation}

 Let us refer to the following  $CT \times CT$ matrix as a reversed FIM:
 \begin{equation}
F^* :=R^{\top}R,
 \end{equation}
 which is the right-to-left reversed Gram matrix of $F$. This $F^*$ is essentially the same as the NTK \cite{jacot2018neural}. 
 The $F$ and $F^*$ have the same non-zero eigenvalues by definition. 
 \citet{karakida2018universal} introduced $F^*$ to derive the eigenvalue statistics in Theorem 2.2.
 Technically speaking, they derived the eigenvalue statistics under the gradient independence assumption (Assumption 3.2). However,  \citet{yang2019scaling} recently succeeded in proving that this assumption is unnecessary. Therefore, Theorem 2.2 is free from this assumption.

To evaluate the effects of batch normalization, we need to take a more careful look into $F^*$ than done in previous studies. As shown in Supplementary Material B, the FIM under  batch normalization in the last layer requires information on how fast backward order parameters asymptotically converge in the large $M$ limit. Let us introduce the following variables depending on $M$: 
\begin{equation}
      \tilde{q}^l_{M,t} := \sum_i \delta_i^l(t)^2, \ \ \tilde{q}^l_{M,st} := \sum_{i}  \delta_i^{l} (s) \delta_{i}^{l}(t). 
\end{equation}
When $C$ is a constant of order $1$, $\delta_i^l(t)$ is of order $1/\sqrt{M}$ and the above summations become of order $1$. The variable $\tilde{q}^l_{M,t}$ is a special case of $\tilde{q}^l_{M,st}$ with $s=t$. 
 Recent studies \cite{yang2019scaling,arora2019exact} proved that, in the large $M$ limit,  backward order parameters asymptotically converge to $\tilde{q}^l_{st}$ satisfying the recurrence relations (Eq. (\ref{eq_tilqst})). Suppose that we have in the large $M$ limit, 
 \begin{equation}
     \tilde{q}^l_{M,st} = \tilde{q}^l_{st} + O(1/M^q),
 \end{equation}
 where $q>0$ determines a convergence rate. 
 \citet{schoenholz2016} derived $  \tilde{q}^l_{\infty,st} = \tilde{q}^l_{st}$ under the gradient independence assumption.
 \citet{yang2019scaling} succeeded in deriving the recurrence relations without using the gradient independence assumption.
  It also gave an upper bound of the residual term $O(1/M^q)$ although an explicit value of $q$ was not shown.
 \citet{arora2019exact} also succeeded in deriving the recurrence relations in ReLU networks by using a non-asymptotic method and obtained $q \geq 1/4$ (Theorem B.2 in \cite{arora2019exact}). 
 Thus, previous studies have paid much attention to $\tilde{q}^l_{st}$ while there has been almost no comprehensive discussion regarding  the residual term $O(1/M^q)$.
 As shown in Supplementary Material B.1.3, we confirmed that $q=1/2$ holds in simulations of typical DNN models and that the gradient independence assumption yields $q=1/2$.

Between the reversed FIM and convergence rate $q$, we found that the following lemma holds. This lemma is a minor extension of Supplementary Material A in \cite{karakida2018universal} into the case without the gradient independence assumption.

\begin{lem}
{\it Suppose a non-centered network and i.i.d. input samples generated by Eq. (\ref{eq7:0410}).
When $M$ is sufficiently large, the $F^*$ can be partitioned into $C^2$ block matrices whose $(k,k')$-th block is a $T \times T$ matrix defined by     
\begin{equation}
  F^{*}(k,k') =   \alpha \frac{M}{T} K \delta_{kk'}+ \frac{1}{T}O(M^{1-q^*}), \label{eq69:0110}
\end{equation}
where $q^*=\min\{q,1/2\}$,  $k,k'=1,...,C$ and  $\delta_{kk'}$ is the Kronecker delta. The matrix $K$ has entries given by
\begin{equation}
K_{st}=   \kappa_1 \ \  (s=t), \ \   \kappa_2 \ \ (s \neq t). 
\end{equation}
}
\end{lem}

{\it Proof.}
We have the parameter set $\theta=\{W^l_{ij} , b^l_i \}$ but the number of bias parameters (of $O(M)$) is much less than that of weight parameters (of $O(M^2)$). Therefore, the contribution of the FIM corresponding the bias terms are negligibly small in the large $M$ limit \cite{karakida2018universal}, and what we should analyze is weight parts of the FIM, that is, $\nabla_{W_{ij}^l}f_k$.
 The $(k,k')$-th block of $F^*$ has the $(s,t)$-th entry as
\begin{align}
F^*(k,k')_{st}&=\sum_l\sum_{ij} \nabla_{W^l_{ij}} f_k^\top (s) \nabla_{W^l_{ij}}  f_{k'} (t)/T \\
&= \sum_l M_{l-1}(\sum_i  \delta_{k,i}^l (s)  \delta_{k',i}^l (t)) \hat{q}^{l-1}_{M,st}/T 
\label{A28t}
\end{align}
for $s,t=1,...,T$. 
In the large $M$ limit, we can apply the central limit theorem to the feedforward propagation because the pre-activation $u_i^l$ is a weighted sum of independent random weights \cite{amari1974method,poole2016}:
$ \hat{q}^{l}_{M,st} = \hat{q}^l_{st}+ O(1/\sqrt{M})$. This convergence rate of $1/2$ is also known in non-asymptotic evaluations \cite{daniely2016toward,arora2019exact}.  We then have 
\begin{align}
     F^*(k,k')_{st}&=  \sum_l M_{l-1}(\tilde{q}^l_{st}\delta_{kk'}+O(1/{M^q})) (\hat{q}^{l-1}_{st}+O(1/\sqrt{M}))/T  \\
     &= \sum_l \frac{M_{l-1}}{T} \tilde{q}^l_{st}\hat{q}^{l-1}_{st} \delta_{kk'} + \frac{1}{T}O({M}^{1-q*}) \\
     &= \alpha \kappa_2\frac{M}{T} \delta_{kk'}+\frac{1}{T}O({M}^{1-q*}) ,\label{eq12:0522}
\end{align}
where $q^*:=\min\{q,1/2\}$. 
The backward order parameters for $k\neq k'$ become zero because the chains do not share the same weight $W^L_{ki}$ and the initialization of the recurrence relations (Eq. (\ref{eq_tilqst})) becomes $\tilde{q}_{st}^L=\sum_i  \delta_{k,i}^L (s)  \delta_{k',i}^L (t)=\sum_i  \delta_{ki} \delta_{k'i}=0$.
When $t=s$, we have $F^*(k,k')_{tt} = \alpha \kappa_1 M/T \delta_{kk'} + O(M^{1-q^*})/T$.
\qed

\com{
The current work essentially differs from  \cite{karakida2018universal} in the point that 
the evaluation of $F^*$ includes the convergence rate. 
The previous work investigated DNNs without any normalization method and such cases allow us to focus on the first term of the right-had side of Eq. (\ref{eq69:0110}). 
This is because the second term becomes asymptotically negligible in the large $M$ limit.
In contrast, batch normalization in the last layer makes the first term comparable to the second term and requires careful evaluation of the second term. Thus, eigenvalues statistics become dependent on the convergence rate. 
}

The previous work \cite{karakida2018universal}
showed that the matrix $K$ in the first term of (\ref{eq69:0110}) 
determines the eigenvalue statistics such as $m_\lambda$ and $\lambda_{max}$ in the large $M$ limit. 
 The assumption of i.i.d. input samples makes the structure of matrix $K$ easy to analyze, i.e., all the diagonal terms take the same $\kappa_1$ and all the  non-diagonal terms take $\kappa_2$.
Using this matrix $K$, we can also derive 
the eigenvectors of $F^*$ corresponding to $\lambda_{max}$: 
 
\begin{lem}[Supplementary Material A.4 in \cite{karakida2018universal}]
{\it Suppose a non-centered network and i.i.d. input samples generated by Eq.  (\ref{eq7:0410}). Denote the eigenvectors of $F^*$ corresponding to $\lambda_{max}$ as  $\nu_k \in \mathbb{R}^{CT}$ $(k=1,...,C)$. In the large $M$ limit, they are asymptotically equivalent to
\begin{align}
 (\nu_k)_i := \begin{cases} 1/\sqrt{T} & ( (k-1)T+1 \leq i \leq kT), \\  
  0 & (\mathrm{otherwise}). \end{cases} \label{eq44:0323}
\end{align}
}
\end{lem}

It should be remarked that the above results require $\kappa_2>0$.  
Technically speaking, the second term of  Eq. (\ref{eq69:0110})  is negligible because $\kappa_1$ is positive by definition and $\kappa_2$ is also positive in a non-centered network.   
If one considers a centered network, however, the initialization of recurrence relations, i.e., $\hat{q}^0_{st}=0$, recursively yields 
\begin{equation}
  {q}^{l+1}_{st}= \sigma_w^2I_{\phi}[q^l_t,0]  = \sigma_w^2  \int Dy Dx \phi (\sqrt{q_t^{l}}x)\phi (\sqrt{q_t^{l}}y )=0,
\end{equation}
 and it gives $\hat{q}^l_{st}=0$ for all $l$ and $\kappa_2=0$. In such cases, the second term of  Eq. (\ref{eq69:0110}) dominates the non-diagonal entries of $F^*(k,k')$ and affects the eigenvalue statistics. In contrast, we have ${q}_{st}^l>0$ and $\hat{q}_{st}^l>0$ in a non-centered network. Because  $\tilde{q}_{st}^l>0$ holds as well, we have $\kappa_2>0$ and the second term becomes negligible in the large $M$ limit.


\subsection{Eigenspace of $\lambda_{max}$}
To prove Theorem 3.1, we use the eigenvector $\nu_k$ obtained in Lemma A.2. 
The eigenspace of $F$ corresponding to $\lambda_{max}$ is constructed from $\nu_k$. Let us denote an eigenvector of $F$ as $v$ satisfying $Fv=\lambda_{max}v$. By multiplying $R^{\top}$ by both sides, we have
\begin{equation}
F^*(R^{\top}v)= \lambda_{max}(R^{\top}v).
\end{equation} This means that $R^{\top}v$ is the eigenvector of $F^*$. Then, we obtain $R^{\top}v=\sum_k c_k \nu_k$ up to a scale factor by using coefficients $c_k$ satisfying $\sum_k c_k=1$.
Substituting it into $(RR^{\top})v=\lambda_{max}v$, we have $v=\sum_k c_k R\nu_k=\sum_k c_k \mathrm{E}[\nabla_\theta f_k]$. 
As a result, the eigenspace of $F$ corresponding to $\lambda_{max}$ is spanned by $\mathrm{E}[\nabla_\theta f_k]$. It is easy to conform that the derivative $\nabla_{W^l_{ij}} f_k (=\delta_i^l h^{l-1}_j)$ is of $O(1/\sqrt{M})$ and we have 
\begin{equation}
F\cdot \mathrm{E}[\nabla_\theta f_k] = \lambda_{max}  \mathrm{E}[\nabla_\theta f_k] + O(M^{1/2-q^*}), \label{eqx:0524}
\end{equation}
for $k=1,...,C$.  The first term of the right-hand side of   
Eq. (\ref{eqx:0524}) is of $O(M^{1/2})$  in non-centered networks and asymptotically larger than the second term.
Thus, we obtain Theorem 3.1.

Note that Lemma A.2 requires non-centered networks and so does Theorem 3.1. Pathological sharpness appears because non-centered networks make the first term of Eq. (\ref{eqx:0524})  non-negligible.
In contrast, centered networks have  $\kappa_2=0$ and the order of the above $\lambda_{max}$ becomes lower for a sufficiently large $T$.
 In such case,  the second term of  
Eq. (\ref{eqx:0524}) becomes dominant and we cannot judge whether $\mathrm{E}[\nabla_\theta f_k]$  is the eigenvector of FIM or not.

\section{Batch normalization in last layer}

\subsection{Mean subtraction}
\subsubsection{Mean of eigenvalues}

The FIM under the mean subtraction (Eq. (\ref{outputNorm})) is expressed by 
\begin{equation}
    F_{L,mBN}= \sum_k \mathrm{E}[ \nabla_\theta \bar{f}_k(t) \nabla_\theta \bar{f}_k(t)^\top]  
    = (R-\bar{R})(R-\bar{R})^\top,
\end{equation}
where $\bar{R}$ is  a $CT \times P$ matrix whose $k$-th column  is given by a vector $\nabla_\theta \mu_i/\sqrt{T}$ ($(i-1)T+1 \leq k \leq iT$, $i=1,2,...,C$). Note that the hyperparameter $\beta_k$ disappears since $\beta_k$ is independent of $\theta$.
Here, we define the projector 
\begin{equation}
    G:= I_{T} - 1_T(1_T)^\top/T,
\end{equation}
which satisfies $G^2=G$. 
Using this projector, we have $RG=R-\bar{R}$ and
\begin{equation}
    F_{L,mBN}=  R(I_{C} \otimes G)R^\top,
\end{equation}
where $I_{C}$ is a $C\times C$ identity matrix and $\otimes$ is the Kronecker product.  
We introduce a reversed Gram matrix of the FIM under the mean subtraction:
\begin{equation}
    F_{L,mBN}^* :=(R-\bar{R})^\top (R-\bar{R})= (I_{C} \otimes G)  F^* (I_{C} \otimes G)^\top.
\end{equation}
Let us partition $ F_{L,mBN}^*$ into  $C^2$ block matrices and denote its $(k,k')$-th block as a $T \times T$ matrix $F^*_{L,mBN}(k,k')$.
 Substituting the $F^*$ (\ref{eq69:0110}) into the above, we obtain these blocks as 
\begin{equation}
  F^*_{L,mBN}(k,k') =   \alpha \frac{M}{T} K_{L,mBN} \delta_{kk'}+ \frac{1}{T}O(M^{1-q^*}), \label{eq37:0510}
\end{equation}
with 
\begin{equation}
(K_{L,mBN})_{st}:=\begin{cases}   (\kappa_1-\kappa_2)(1-1/T)  & (s=t), \\   -(\kappa_1-\kappa_2)/T & (s \neq t).
\end{cases}
\end{equation}
We assume $T \geq 2$ since $T=1$ is trivial.

The mean of eigenvalues is asymptotically obtained by 
\begin{align}
     m_\lambda &= \mathrm{Trace}(F_{L,mBN}^*)/P  \\ 
     &\sim C(1-1/T)(\kappa_1-\kappa_2)/M.
\end{align}

\subsubsection{Largest eigenvalue when $T\geq 2$ and $T=O(1)$}

First, we obtain a lower bound of $\lambda_{max}$.
In general, we have 
\begin{equation}
\lambda_{max} = \max_{{v}; ||{v}||^2=1} {v}^\top F^{*}_{L,mBN} {v}.
\end{equation}
 We then find
\begin{align}
\lambda_{max} &\geq {\nu}^\top F^{*}_{L,mBN} {\nu},  \label{eqE6}
\end{align}
where $\nu$ is a $CT$-dimensional vector whose $((i-1)T+1)$-th entries are  $1/\sqrt{2C}$,  $((i-1)T+2)$-th entries are $-1/\sqrt{2C}$, and the others are $0$ ($i=1,...,C$). We then  have 
\begin{equation}
    \lambda_{max} \geq \alpha \frac{\kappa_1-\kappa_2}{T}M + O(M^{1-q^*}). \label{eq46:0518}
\end{equation}

Next, we obtain an upper bound of $\lambda_{max}$. 
In general,  the maximum eigenvalue is denoted as the spectral norm $||\cdot||_2$, i.e.,  $\lambda_{max}=||F^*_{L,mBN}||_2$.
 Using the triangle inequality, we have
\begin{equation}
\lambda_{max} \leq ||\bar{F}^*_{L,mBN}||_2+ ||\tilde{F}^*_{L,mBN}||_2.
\end{equation}
We divided ${F}^*_{L,mBN}$ into $\bar{F}^*_{L,mBN}$ corresponding to the first term of Eq. (\ref{eq37:0510}) and $\tilde{F}^*_{L,mBN}$ corresponding to the second term.
$\bar{F}^*_{L,mBN}$ is composed of $\alpha \frac{M}{T} K_{L,mBN}$.  The eigenvalues of  $\alpha \frac{M}{T} K_{L,mBN}$ are explicitly obtained as follows: $\lambda_{1}=  0$ for an eigenvector $e=(1,...,1)$, and $\lambda_i= \alpha  (\kappa_1 -\kappa_2)M/T $  for eigenvectors $e_1 -e_i$ ($i=2,...,T$), where $e_i$ denotes a unit vector whose entries are $1$ for the $i$-th entry and $0$ otherwise. We then obtain
\begin{equation}
||\bar{F}^*_{L,mBN}||_2= \alpha  (\kappa_1 -\kappa_2)M/T.
\end{equation}
Each entry of $\tilde{F}^*_{L,mBN}$ is of $O(M^{1-q^*})$. We then have 
\begin{equation}
 ||\tilde{F}^*_{L,mBN}||_2 \leq  ||\tilde{F}^*_{L,mBN}||_F=O(M^{1-q^*}),   
\end{equation}
where $||\cdot||_F$ is the Frobenious norm. These lead to 
\begin{equation}
\lambda_{max} \leq \alpha  \frac{\kappa_1 -\kappa_2}{T}M+ O(M^{1-q^*}). \label{eq54:0518}
\end{equation}

Finally, sandwiching $\lambda_{max}$ by bounds (\ref{eq46:0518}) and (\ref{eq54:0518}), we asymptotically obtain  
\begin{equation}
\lambda_{max} \sim \alpha  \frac{\kappa_1 -\kappa_2}{T}M,
\end{equation}
in the large $M$ limit. 

Note that $\kappa_1>\kappa_2$ holds in our settings. 
We can easily observe $\hat{q}^{l}_{t}>\hat{q}^{l}_{st}$ from the Cauchy–Schwarz inequality and it leads to $\kappa_1>\kappa_2$ (strictly speaking, when $\phi(x)$ is a constant function,  its equality holds and we have $\hat{q}^{l}_{t}=\hat{q}^{l}_{st}$ and $\kappa_1=\kappa_2$. However, we do not suppose the constant function as an ''activation'' function and then $\kappa_1>\kappa_2$ holds).

\subsubsection{Largest eigenvalue when $M/T=const.$}

This case requires a careful consideration of the $O(M^{1-q^*})$ term in the reversed FIM (\ref{eq37:0510}). This is because the non-diagonal term of $K_{L,mBN}$ asymptotically decreases to zero in the large $M$ limit and the $O(M^{1-q^*})$ term becomes non-negligible.
We found the following theorem without using the gradient independence assumption; 

\begin{thm}
{\it  Suppose a non-centered network with the mean subtraction in the last layer (Eq. (\ref{outputNorm})) and i.i.d. input samples generated by Eq. (\ref{eq7:0410}). When $T=O(M)$ with a constant $\rho:=M/T$, the largest eigenvalue in the large $M$ limit is asymptotically evaluated as 
\begin{equation}
   \rho \alpha (\kappa_1 - \kappa_2) +  c_1  M^{1-2q^*} \leq  \lambda_{max}  \leq \sqrt{ C\alpha^2 \rho (\kappa_1 - \kappa_2)^2M + c_2 M^{2(1-q^*)} }
\end{equation}
for $q^* = \min \{q,1/2\}$, non-negative constants $ c_1$ and $c_2$. 
}
\end{thm}

{\it Proof.} \ \ 
To evaluate the largest eigenvalue, we use the second moment of the eigenvalues, i.e., $s_\lambda:= \sum_i^P \lambda_i^2/P$. 
Because $F^*_{L,mBN}$ is positive semi-definite,  we have $\sum_i \lambda_i^2 = \sum_{st}((F^*_{L,mBN})_{st})^2$ and obtain 
\begin{equation}
    s_\lambda = \sum_{k,k'} \sum_{s,t} \left( \alpha \frac{M}{T} (K_{L,mBN})_{st} \delta_{kk'}+ \frac{1}{T}(F_0(k,k'))_{st} \right)^2/P,
\end{equation}
where we denote the second term of Eq. (\ref{eq37:0510}) as $F_0 = O(M^{1-q^*})$.  We then have 
\begin{align}
    s_\lambda &= \nonumber \\
    \sum_{s,t}  &\left \{ C  \left( \alpha \frac{M}{T} (K_{L,mBN})_{st} \right )^2 + 2 \alpha \frac{M}{T^2} (K_{L,mBN})_{st}  \sum_k (F_0(k,k))_{st} + \sum_{k,k'} \frac{1}{T^2}(F_0(k,k'))_{st}^2 \right \}/P \nonumber \\ 
    &= \frac{C\alpha}{T}(1-1/T) (\kappa_1-\kappa_2)^2 + \frac{2(\kappa_1-\kappa_2)}{MT^2}(1-1/T) \sum_{t}  (K_{L,mBN})_{tt}  \sum_k (F_0(k,k))_{tt} \nonumber \\ 
    &-\frac{2(\kappa_1-\kappa_2)}{MT^3} \sum_{s \neq t}  (K_{L,mBN})_{st}  \sum_k (F_0(k,k))_{st} + \sum_{s,t} \sum_{k,k'} \frac{1}{\alpha T^2M^2}(F_0(k,k'))_{st}^2. \label{eq32:0521}
\end{align}
When $T=O(M^p)$ ($p\geq 0$), the first term is of $O(1/M^p)$, the second and third terms are of $O(1/M^{p+q^*})$, and the fourth term is of $O(1/M^{2q^*})$. 
Therefore, the second and third terms are negligible compared to the first term for all $p$ and $q^*$. The fifth term is non-negative by definition. 
Although we can make the bounds of $\lambda_{max}$ for all $p$, we focus on $p=1$ for simplicity. In the large $M$ limit, we have asymptotically
\begin{equation}
s_\lambda \sim \frac{C\alpha\rho}{M} (\kappa_1-\kappa_2)^2 +\frac{c_0}{ M^{2q^*}}.
\end{equation}
The constant $c_0$ comes from the fourth term of Eq. (\ref{eq32:0521}) and is non-negative.

The lower bound of $\lambda_{max}$ is derived from $\lambda_{max} \geq \sum_i \lambda_i^2/\sum_i \lambda_i = s_\lambda/m_\lambda$, that is, 
\begin{align}
      \lambda_{max}  &\geq  (\kappa_1 - \kappa_2)\rho +  c_1  M^{1-2q^*}.
\end{align}
The upper bound comes from $\lambda_{max} \leq \sqrt{\sum_i \lambda_i^2}=\sqrt{P s_\lambda}$ and we have 
\begin{equation}
\lambda_{max}  \leq \sqrt{ C\alpha^2 \rho (\kappa_1 - \kappa_2)^2M + c_2 M^{2(1-q^*)} }.
\end{equation}
The non-negative constants $c_1$ and $c_2$ come from $c_0$.
\qed

Thus, we find that $\lambda_{max}$ is of order $M^{1-2q^*}$ at least and  of order $M^{1-q^*}$ at most. Since we have $0<q^*\leq 1/2$ by definition,  the order of $\lambda_{max}$ is always  lower than
order of $M$. Therefore, we can conclude that the mean subtraction alleviates the pathological sharpness for any $q$.

\renewcommand{\thefigure}{S.1}
\begin{figure}
\vskip 0.2in
\begin{center}
\centerline{\includegraphics[width=0.8\columnwidth]{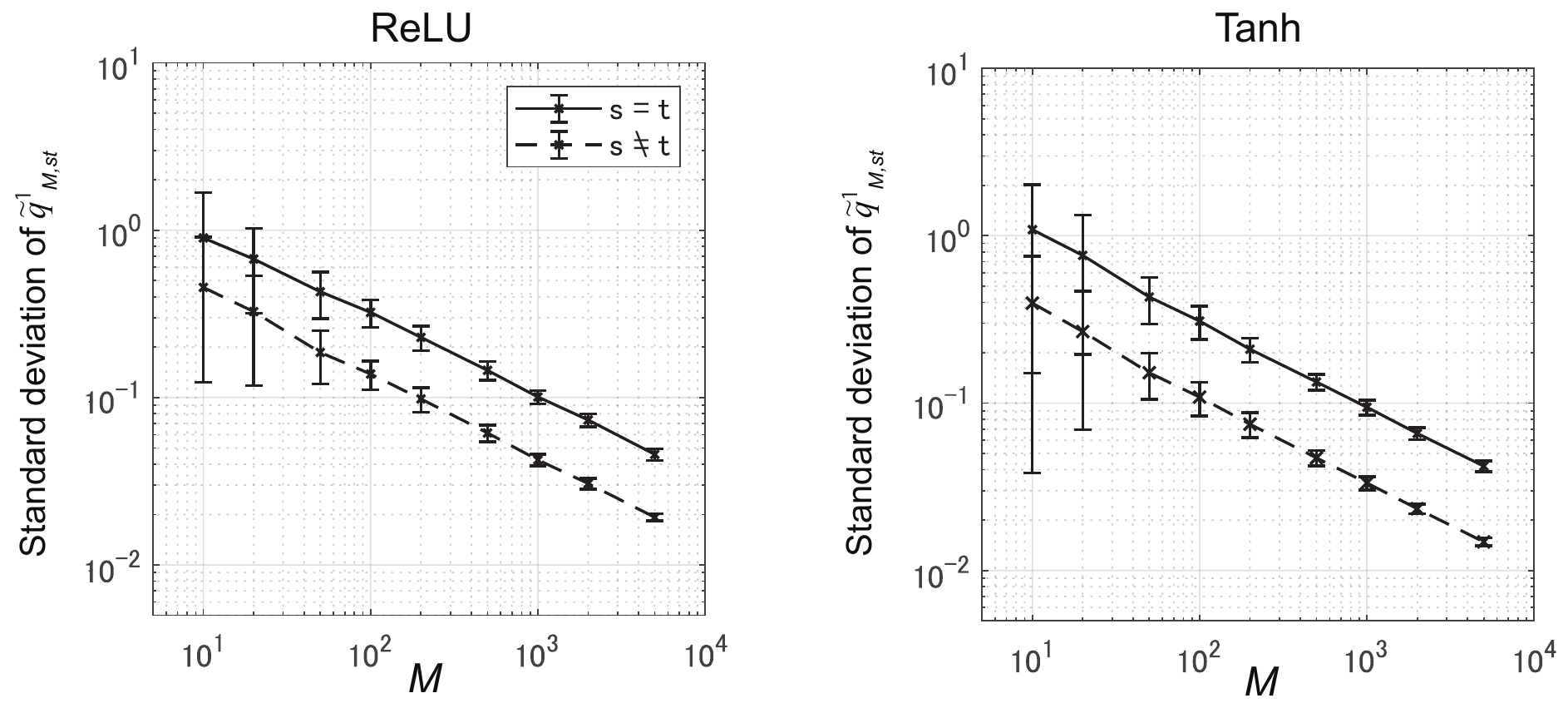}}
\caption{Standard deviation of $\tilde{q}_{M,st}^1$.  Variances of parameters were given by  ($2,0$) in ReLU case and ($\sigma_w^2,\sigma_b^2$) = ($3,0.64$) in tanh case.  
}
\end{center}
\vskip -0.2in
\end{figure}

Furthermore, we confirmed that $q=q^*=1/2$ held in simulations as shown in Fig. S.1.  We numerically computed $\tilde{q}^1_{M,st}$ in DNNs with  input samples generated by Eq. (\ref{eq7:0410}), random Gaussian weights and biases. We set $\alpha_l=C=1$, $L=3$ and $T=100$. 
The experiments involved $100$ different ensembles.
We observed that the standard deviation of $\tilde{q}^1_{M,st}$ decreased with the order of $1/\sqrt{M}$, which indicated that $q=1/2$ held in the real models.

From the theoretical perspective, we found that the gradient independence assumption achieves $q=q^*=1/2$ and leads to a constant lower bound independent of $M$. \begin{lem}
{\it The gradient independent assumption yields  $q=q^*=1/2$.
}
\end{lem}
{\it Proof.} 
 Under the gradient assumption, we can apply the central limit theorem to  $\tilde{q}^{l}_{M,st}$ because we can assume that they do not share the same random weights and biases. That is, 
 \begin{align}
    \tilde{q}_{M,st}^l &= \sum_i   \phi'^{l}_i(s) \phi'^{l}_{i}(t)  \sum_{j,j'} W^{l+1}_{ji} W^{l+1}_{j'i} \delta_j^{l+1}(s) \delta_{j'}^{l+1}(t) \\ 
    &=     \sum_{i}  \phi'^{l}_i(s) \phi'^{l}_{i}(t) \sum_{j,j'} \tilde{W}^{l+1}_{ji} \tilde{W}^{l+1}_{j'i} \delta_j^{l+1}(s) \delta_{j'}^{l+1}(t)  \ \ \ \  (\text{by Assumption 3.2}) \\
    &=   \sum_{i}  \phi'^{l}_i(s) \phi'^{l}_{i}(t) \frac{\sigma_w^2}{M_l} \sum_{j} \delta_j^{l+1}(s) \delta_{j}^{l+1}(t) + O(1/\sqrt{M})  \ \ \ \  (\text{in the large $M$ limit}) \\ 
    &= \sigma_w^2 \tilde{q}_{st}^{l+1}  I_{\phi'}[q^l_t,q_{st}^l]+ O(1/\sqrt{M}).
\end{align}
 Thus, we have $\tilde{q}^l_{M,st} = \tilde{q}^l_{st} + O(1/\sqrt{M})$ and obtain   $q=q^*=1/2$. 
\qed

The bounds for $\lambda_{max}$ in Theorem 3.3 are immediately obtained from Theorem B.1 and Lemma B.2.

\subsection{Mean subtraction and variance normalization}

Define $\bar{u}_k(t)=:u_k^L (t) - \mu_k(\theta)$.
The derivatives of output units are given by 
\begin{align}
    \nabla_\theta f_k &= \frac{1}{\sigma_k(\theta)} \nabla_\theta \bar{u}_k - \frac{1}{\sigma_k(\theta)^3} \bar{u}_k  \mathrm{E}[ \bar{u}_k \nabla_\theta \bar{u}_k ]. \label{eq55:0517}
\end{align}
Then,  the FIM is given by
\begin{align}
    F_{L,BN} &:= \sum_k^C \mathrm{E}[ \nabla_\theta f_k(t) \nabla_\theta f_k(t)^\top] \\ 
    &=  \sum_k^C \frac{1}{\sigma_k(\theta)^2} \left( \mathrm{E}[\nabla_\theta \bar{u}_k \nabla_\theta \bar{u}_k^\top] - \frac{1}{\sigma_k(\theta)^{2}}\mathrm{E}[ \bar{u}_k \nabla_\theta \bar{u}_k]\mathrm{E}[ \bar{u}_k \nabla_\theta \bar{u}_k]^\top \right),
\end{align}
 using the fact $\sigma_k(\theta)^2=\mathrm{E}[\bar{u}_k^2]$.
We can represent $F_{L,BN}$ in a matrix representation as
\begin{equation}
 F_{L,BN} = (R-\bar{R})Q(R-\bar{R})^\top.
\end{equation}
 $Q$ is a $CT \times CT$ matrix whose $(k,k')$-th block is given by a $T \times T$ matrix,  
\begin{equation}
    Q(k,k'):= \frac{1}{\sigma^2_k}  \left(I_{T}- \frac{1}{T\sigma^{2}_k} \bar{u}_k \bar{u}_k^\top \right) \delta_{kk'}, \label{S47:0524}
\end{equation}
where $I_{T}$ is a $T\times T$ identity matrix, $\sigma_k^2$ means $\sigma_k(\theta)^2$,  and $Q(k,k)$ is a projector to the vector $\bar{u}_k$. 
$F_{L,BN}$ and the following matrix have the same non-zero eigenvalues, 
\begin{align}
    F^*_{L,BN} &:=Q(R-\bar{R})^\top(R-\bar{R}) = Q F^*_{L,mBN}.
\end{align}
$F^*_{L,BN}$ is a $CT \times CT$ matrix and partitioned into $C^2$ block matrices.
Using Eq. (\ref{eq37:0510}), we obtain the $(k,k')$-th block as
\begin{align}
F^*_{L,BN}(k,k') &=\alpha \frac{M}{T}Q(k,k')K_{L,mBN}\delta_{kk'}+\frac{1}{T}O(M^{1-q^*}),
\end{align}
where the independence assumption yields $q^*=1/2$. 
The first term is easy to evaluate, 
\begin{equation}
Q(k,k)K_{L,mBN} =\frac{1}{\sigma^2_k} (\kappa_1-\kappa_2) \left(I_T - \frac{1}{T\sigma^{2}_k}(1-\frac{1}{T})\bar{u}_k \bar{u}_k^\top \right),
\end{equation}
by using the fact of $\sum_t \bar{u}_k^L(t)=0$. Suppose the case of $\rho=M/T=const.$ Regarding the diagonal entries of $Q(k,k')K_{L,mBN}$, the contribution of $\frac{1}{\sigma^{2}_k T}(1-\frac{1}{T})u_k u_k^\top $ is negligible to that of  $I_T$ in the large $T$ limit.
Thus, we asymptotically obtain
\begin{align}
m_\lambda &\sim  \sum_k \frac{1}{\sigma^2_k} (\kappa_1-\kappa_2)/M.
\end{align}
The bounds of the largest eigenvalue are straightforwardly obtained from the second moment as in the deviation of Theorem B.1. Since the second moment $s_\lambda=\sum_i\lambda_i^2/P$ is given by a trace of the squared matrix in general, we have  
\begin{align}
    s_\lambda &= \mathrm{Trace}({F^*_{L,BN}}^2)/P \\
    &= \sum_{k,k'}  \mathrm{Trace}(F^*_{L,BN}(k,k') F^*_{L,BN}(k',k) )/P \\
    &= \alpha^2 \rho^2 \sum_{k}  \mathrm{Trace}(Q(k,k)K_{L,mBN}Q(k,k)K_{L,mBN})/P+O(1/{M}) \\ 
    &= \alpha \rho \sum_k \frac{1}{\sigma_k^4} (\kappa_1-\kappa_2)^2/M +O(1/M).
\end{align}
The lower bound is given by $\lambda_{max} \geq s_\lambda/m_\lambda$ and the upper bound by 
 $\lambda_{max} \leq \sqrt{Ps_\lambda}$.

\section{Batch normalization in middle layers}

Batch normalization makes the chain of backward signals more complicated as follows. Suppose the $t$-th input sample is given. Then, the activation in each layer depends not only on the $t$-th sample but also on the whole of all samples. This is because batch normalization includes $\mu^l$ and $\sigma^l$,  which depend on the whole of all samples in the batch. 
Therefore, we should compute derivatives as 
\begin{equation}
\frac{\partial u^L_k(t)}{\partial W_{ij}^l} = \sum_{a} \delta_{k,i}^l(t;a) h^{l-1}_j(a), \end{equation}
where we defined 
\begin{equation}
    \delta_{k,i}^l(t;a):=\frac{\partial u^L_k(t)}{\partial u_i^l(a)}.
\end{equation}
Its chain rule is given by
\begin{align}
 \delta_{k,i}^l(t;a)  
&= 
 \sum_{b,j}\frac{\partial u_j^{l+1}(b)}{\partial u_i^l(a)}  \delta_{k,j}^{l+1}(t;b)     \\  
 &= \frac{1}{\sigma^l_i} \sum_{b} 
  \phi'^{l}_i(b) P_{i}^l(a,b) \sum_j W^{l+1}_{ji}\delta_{k,j}^{l+1}(t;b),
\end{align}
where we defined 
\begin{equation}
     P_{i}^l(a,b) := \delta_{ab}- \frac{1}{T}-\frac{(u_i^l(b)-\mu_i^l)(u_i^l(a)-\mu_i^l)}{(\sigma^l_i)^2 T}.
\end{equation}
Recently, \citet{yang2019bn} investigated a gradient explosion of the above chain rule in extremely deep networks although it requires a complicated formulation of mean field equations and is analytically intractable in general cases. In the following, we demonstrate an approach to batch normalization in the middle layers by avoiding the complicated analysis of the chain rule.

\subsection{Effect of un-normalized last layer on the FIM}

The derivative with respect to the $L$-th layer is independent of the complicated chain of batch normalization because we do not normalize the last layer and have 
\begin{align}
\frac{\partial f_k(t)}{\partial W_{ij}^L} = \sum_{a} \delta^L_{k,i}(t;a) h^{L-1}_j(a)  = h^{L-1}_j(t) \delta_{ki},
\end{align}
where we used $\delta^L_{k,i}(t;a)=\delta_{ki}\delta_{ta}$. 
The lower bound of $\lambda_{max}$ is derived as follows: 
\begin{align}
\lambda_{max} &= \max_{||x||^2=1;x \in \mathbb{R}^P}  x^\top F x \\
& \geq \max_{||x||^2=1;x \in \mathbb{R}^{CM_{L-1}}}  x^\top F_L x \\ 
&=: \lambda_{max}^L,
\end{align}
where we denote a diagonal block of $F$ as $F_L:=\sum_k \mathrm{E}[\nabla_{\theta^L} f_k \nabla_{\theta^L} f_k^\top]$ and $\theta^L$ is a vector composed of all entries of $W^L$. We denote its largest eigenvalue as $\lambda_{max}^L$. One can represent $F_L=RR^{\top}$ by using    
 $R := 
[\nabla_{\theta^L} f_1 \ \ \nabla_{\theta^L} f_2 \ \  \cdots \ \  \nabla_{\theta^L} f_C]/\sqrt{T}$.
Its reversed FIM is given by $F^*_L :=R^{\top}R$.  In the large $M$ limit, we have 
\begin{equation}
F^{*}_L(k,k') =   \alpha \frac{M}{T} K_L \delta_{kk'}+ \frac{1}{T}O(M^{1-q^*}). 
\end{equation}
The matrix $K_L$ is defined by  
$(K_L)_{st}:=   \hat{q}^{L-1}_{t,BN} \ \  (s=t), \ \   \hat{q}^{L-1}_{st,BN} \ \ (s \neq t) $ where we denote feedroward order parameters for batch normalization as 
\begin{equation}
\hat{q}^{l}_{t,BN} := \frac{\sum_i}{M_{l}} \phi (\bar{u}_i^{l}(t))^2, \ \ \hat{q}^{l}_{st,BN} := \frac{\sum_i}{M_{l}} \phi (\bar{u}_i^{l}(t))\phi (\bar{u}_i^{l}(s)). \label{eq66:0523}
\end{equation}
We then have 
\begin{equation}
\lambda_{max}^L \geq \nu_k^\top F_L \nu_k =   \frac{T-1}{T}\hat{q}^{L-1}_{st,BN} + \frac{\hat{q}^{L-1}_{t,BN}}{T}.
\end{equation}
The evaluation of the order parameters are shown in the following subsection. When the activation function is non-negative, 
the order paramters are positive.  In particular, they are analytically tractable in ReLU networks.

\subsection{Specific values of $ \hat{q}^{L-1}_{t,BN}$ and $ \hat{q}^{L-1}_{st,BN}$}

Order parameters for batch normalization in the middle layers (\ref{eq66:0523}) require a careful integral over a $T$-dimensional Gaussian distribution \cite{yang2019bn}. This is because the pre-activation $\bar{u}_i^l$ depends on all of $u_i^l(t)$ $(t=1,...,T)$ which share the same weight $W_{ij}^l$. Therefore, we generally need 
the integration of $\phi(\bar{u}_i^l(t))$ over the $T$-dimensional Gaussian distribution, that is, 
\begin{equation}
    \hat{q}^l_{t,BN} = \int D {u}^l \phi \left( \frac{{u}^l(t)-\sum_{t'}{u}^l(t')/T}{\sum_t({u}^l(t)-\sum_{t'}{u}^l(t')/T)^2/T} \right)^2, 
\end{equation}
\begin{equation}
    \hat{q}^l_{st,BN} = \int D {u}^l \phi \left( \frac{{u}^l(t)-\sum_{t'}{u}^l(t')/T}{\sum_t({u}^l(t)-\sum_{t'}{u}^l(t')/T)^2/T} \right)\phi \left( \frac{{u}^l(s)-\sum_{t'}{u}^l(t')/T}{\sum_t({u}^l(t)-\sum_{t'}{u}^l(t')/T)^2/T} \right), 
\end{equation}
where  ${u}^l=({u}^l(1),{u}^l(2),...,{u}^l(T))$ is a $T$ dimensional vector and  ${u}^l \sim \mathcal{N}(0,\sigma_w^2 \Sigma_{l-1})$. The $T \times T$ covariance matrix is defined by $(\Sigma_{l-1})_{st}=\hat{q}^{l-1}_{st,BN}$ ($s \neq t$), $\hat{q}^{l-1}_{t,BN}$ ($s=t$).  
These order parameters are positive when the activation function is non-negative (strictly speaking, non-negative and $\phi(x)>0$ for certain $x$).

Although the above integral is analytically intractable in many activation functions,  \citet{yang2019bn} gave profound insight into the integral. For instance,  
Corollary F.10 in \cite{yang2019bn} revealed that the ReLU activation is more tractable, and we have
\begin{equation}
  \hat{q}^{l}_t=1/2, \ \   \hat{q}^l_{st}= \frac{1}{2} J(-1/(T-1)),
\end{equation}
where $J(x):=(\sqrt{1-x^2}+(\pi-\arccos(x))x)/\pi$ is known as the arccosine kernel. 
\citet{wei2019meanfield} proposed a mean field approximation on the computation of order parameters for batch normalization, which is consistent with the above order parameters in the large $T$ limit. 
The previous study \cite{yang2019bn} also proposed some methods to evaluate the order parameters in more general activation functions.

\section{Additional experiment on gradient descent training}

\renewcommand{\thefigure}{S.2}
\begin{figure}[h]
\vskip 0.2in
\begin{center}
\centerline{\includegraphics[width=0.8\columnwidth]{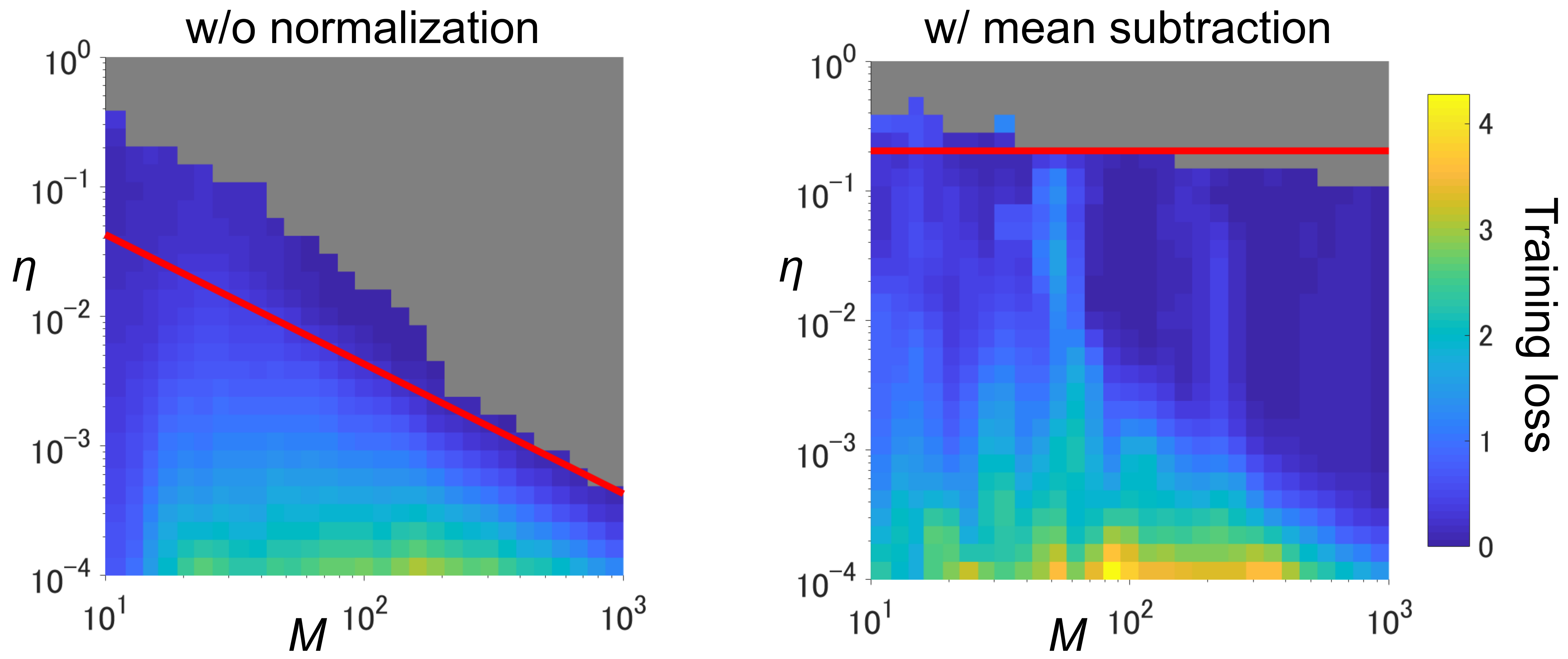}}
\caption{Exhaustively searched training losses depending on $M$ (width) and $\eta$ (learning rate). 
We trained DNNs over ten trials with different random seeds, and plotted the minimum value of the training loss over all trials. 
Gray area corresponds to the explosion of the minimum value (i.e., larger than $10^3$) and means that gradient dynamics in all trials exploded in that area. 
The other experimental settings are the same as those in Fig.2.
The theoretical line predicted well the experimental results of wide networks.
Mean subtraction achieved the larger area of low training losses.  }
\end{center}
\vskip -0.2in
\end{figure}


\section{Layer normalization}

\subsection{Order parameters in layer normalization}

We show that the order parameters under layer normalization are quite similar to those without the normalization. This is because the random weights and biases make the contribution of  layer normalization relatively easy. In the large $M$ limit, we asymptotically have 
\begin{align}
    \mu^l(t) &= \sum_j \left(\frac{\sum_i}{M_l}W_{ij}^l\right) h_j^{l-1}(t)  + \frac{\sum_i}{M_l}b_{i}^l =0,
\end{align}
and 
\begin{align}
    \sigma^l(t)^2 &= \sum_{j,j'} \frac{\sum_i}{M_l}W_{ij}^lW_{ij'}^l h_j^{l-1}(t) h_{j'}^{l-1}(t) +\frac{\sum_i}{M_l}(b_{i}^l)^2  = \sigma_w^2 \hat{q}^{l-1}_t+\sigma_b^2
\end{align}
for $l=1,...,L-1$. 
Let us denote feedforward order parameters as 
\begin{equation}
\hat{q}^{l}_{t} := \frac{\sum_i}{M_{l}} \phi (\bar{u}_i^{l}(t))^2, \ \ \hat{q}^{l}_{st} := \frac{\sum_i}{M_{l}} \phi (\bar{u}_i^{l}(t))\phi (\bar{u}_i^{l}(s)).
\end{equation}
The same calculation as in the feedforward propagation without normalization leads to
\begin{equation}
  \hat{q}^{l+1}_t = \int Du\phi^2 \left( u \right), \ \  \hat{q}^{l+1}_{st} =  I_{\phi}\left[1,\frac{\sigma_w^2 \hat{q}_{st}^{l}+\sigma_b^2}{\sigma_w^2\hat{q}^{l}_t+\sigma_b^2}\right]. \label{eq79:0517}
\end{equation}

The backward order parameters are also very similar to those without layer normalization.  
Let us consider the chain rule which appears in a FIM:  
\begin{equation}
\frac{\partial u_k^L(t)}{\partial W_{ij}^l} = \delta_{k,i}^l(t) h^{l-1}_j(t). 
\end{equation}
 Ommiting index $k$ in $\delta_{k,i}^l(t)$ to avoid complicated notation, we have 
\begin{align}
 \delta_i^l(t) &= 
 \sum_{j}\frac{\partial u_j^{l+1}}{\partial u_i^l}(t)  \delta_j^{l+1}(t)     \\
 &= \frac{1 }{\sigma^l(t)} 
  \sum_k \phi'^{l}_k(t) P_{ki}^l(t) \sum_j W^{l+1}_{jk}\delta_j^{l+1}(t),
\end{align}
where we define $P_{ki}^l(t):=\frac{\partial \bar{u}^l_k}{\partial {u}^l_i}(t)$, which is an essential effect of layer normalization on the chain, and it becomes
\begin{equation}
     P_{ki}^l(t) = \delta_{ki}-\frac{1}{M_l}-\frac{n^l_k(t)n^l_i(t)}{M_l},
\end{equation}
where we define $n^l_k(t):=(u_k^l(t)-\mu^l(t))/\sigma^l(t)$. 
Let us denote backward order parameters as $ \tilde{q}^l_{t} := \sum_i \delta_i^l(t)^2$ and $\tilde{q}^l_{st} := \sum_{i}  \delta_i^{l} (s) \delta_{i}^{l}(t)$ in the large $M$ limit.  We then have 
\begin{align}
    \hat{q}_{st}^l &= \frac{1}{\sigma^l(t)\sigma^l(s)} \sum_i   \sum_{k,k'} \phi'^{l}_k(s) \phi'^{l}_{k'}(t) P_{ki}^l(s) P_{k'i}^l(t) \sum_{j,j'} W^{l+1}_{jk} W^{l+1}_{j'k'} \delta_j^{l+1}(s) \delta_{j'}^{l+1}(t) \\ 
    &=    \frac{1}{\sigma_w^2 \hat{q}^{l-1}_t+\sigma_b^2} \sum_{k,k'}  \phi'^{l}_k(s) \phi'^{l}_{k'}(t) \Gamma_{k,k'}^l(s,t)  \sum_{j,j'} W^{l+1}_{jk} W^{l+1}_{j'k'} \delta_j^{l+1}(s) \delta_{j'}^{l+1}(t),\label{eq80:0523}
\end{align}
where we substituted $\sigma^l(t)=\sigma^l(s)= \sqrt{\sigma_w^2 \hat{q}^{l-1}_t+\sigma_b^2}$ and defined 
\begin{equation}
   \Gamma_{k,k'}^l(s,t) :=\delta_{kk'}-\frac{n^l_k(t)n^l_{k'}(t)+n^l_k(s)n^l_{k'}(s)+1}{M_l}-\frac{n^l_k(s)n^l_{k'}(t) \sum_i n^l_i(s)n^l_{i}(t)}{M_l^2}. 
\end{equation}
Under the gradient independence assumption, we can replace $W_{jk}^{l+1}$ and $W_{j'k'}^{l+1}$  in Eq. (\ref{eq80:0523}) with  $\tilde{W}_{jk}^{l+1}$ and $\tilde{W}_{j'k'}^{l+1}$ which are freshly generated from $\mathcal{N}(0,\sigma_w^2/M_l)$. This is a usual trick in mean field theory of DNNs \cite{schoenholz2016,yang2017,xiao2018dynamical,yang2019bn}. In the large $M$ limit, we have 
\begin{align}
 \tilde{q}_{st}^l 
        &= \frac{1}{\sigma_w^2\hat{q}^{l-1}_t+\sigma_b^2} \sum_{k} \phi'^{l}_k (s)  \phi'^{l}_k(t) \Gamma_{k,k}^l(s,t) \frac{\sigma_w^2}{M_l} \tilde{q}_{st}^{l+1}, \\
\end{align}
where
\begin{equation}
       \Gamma_{k,k}^l(s,t) =1-\frac{n^l_k(t)^2+n^l_k(s)^2+1}{M_l}-\frac{n^l_k(s) n^l_k(t) \sum_i n^l_i(s)n^l_{i}(t)}{M_l^2}.
\end{equation}
The first term of $ \Gamma_{k,k}^l(s,t)$ is dominant in the large $M$ limit because other terms are of order $1/M$. Then, we have    
\begin{align}
 \tilde{q}_{st}^l 
        &= \frac{1}{\sigma_w^2\hat{q}^{l-1}_t+\sigma_b^2} \sum_{k} \phi'^{l}_k (s)  \phi'^{l}_k(t)  \frac{\sigma_w^2}{M_l} \tilde{q}_{t}^{l+1}. \\
\end{align}
After applying the central limit theorem to $\sum_{k}\frac{\phi'^{l}_k (s)  \phi'^{l}_k(t) }{M_l}$, we have 
\begin{equation}
  \tilde{q}^{l}_t = \frac{\sigma_w^2\tilde{q}^{l+1}_t}{\sigma_w^2\hat{q}^{l-1}_t+\sigma_b^2} \int Du [\phi' \left( u \right)]^2, \ \  \tilde{q}^{l}_{st} = \frac{\sigma_w^2\tilde{q}^{l+1}_{st}}{\sigma_w^2\hat{q}^{l-1}_t+\sigma_b^2} I_{\phi'}\left[1,\frac{\sigma_w^2 \hat{q}_{st}^{l-1}+\sigma_b^2}{\sigma_w^2 \hat{q}^{l-1}_t+\sigma_b^2}\right]. \label{eq89:0517}
\end{equation}

 \subsection{FIM}
\subsubsection{Effect of the normalization in the last layer}
Denote the mean subtraction in the last layer as $\bar{u}_k(t)=:u_k^L (t) - \mu^L(t)$. 
The derivatives in the last layer are given by 
\begin{align}
    \nabla_\theta f_k(t) &= \frac{1}{\sigma(t)}\nabla_\theta \bar{u}_k(t)  - \frac{1}{C\sigma(t)^{3}}  \bar{u}_k(t) \sum_i \bar{u}_i(t) \nabla_\theta  \bar{u}_i(t), \label{eq86:0517}
\end{align}
where  $\sigma(t)^2:= \sum_k \bar{u}_k(t)^2 /C$.
Then, the FIM is given by 
\begin{align}
    F_{LN} &:= \sum_k \mathrm{E}[ \nabla_\theta f_k(t) \nabla_\theta f_k(t)^\top] \\ 
    &=\mathrm{E}\left[\frac{1}{\sigma(t)^2}\sum_k \nabla_\theta \bar{u}_k(t)\nabla_\theta \bar{u}_k(t)^\top -\frac{1}{C\sigma(t)^4}\sum_{k,k'} \bar{u}_{k}(t)\bar{u}_{k'}(t) \nabla_\theta \bar{u}_k(t)  \nabla_\theta \bar{u}_{k'}(t)^\top  \right].
\end{align}

We can represent $F_{L,BN}$ in a matrix representation. Define a $P\times CT$ matrix $R$ by
\begin{eqnarray}
R &:=& 
\frac{1}{\sqrt{T}}[\nabla_\theta u_1^L \ \ \nabla_\theta u_2^L \ \  \cdots \ \  \nabla_\theta u_C^L]. 
\end{eqnarray}
Its columns are the gradients on each input sample, i.e., $\nabla_{\theta} {u}_k^L(t)$ $(t=1,...,T)$. 
We then have
\begin{equation}
 F_{LN} = (R-\bar{R})Q(R-\bar{R})^\top,
\end{equation}
where $\bar{R}$ is defined as  a $CT \times P$ matrix whose $((k-1)T+t)$-th column  is given by a vector $\nabla_\theta \mu^L(t)$ ($t=1,...,T$, $k=1,...,C$). We also defined a $CT \times CT$ matrix $Q$ whose $(k,k')$-th block matrix is given by the following $T \times T$ matrix:
\begin{equation}
    Q(k,k')_{st}:=  \frac{1}{\sigma(t)^2} \left( \delta_{kk'} - \frac{1}{C\sigma(t)^2}  \bar{u}_k^L(t) \bar{u}_{k'}^L(t) \right)\delta_{st}
\end{equation}
for $k,k'=1,...,C$. This $Q(k,k')$ is a diagonal matrix. Compared to the matrix $Q$ in  batch normalization (Eq. (\ref{S47:0524})), $Q$ in layer normalization is not block-diagonal. This is because layer normalization in the last layer yields interaction between different output units.  

We introduce the following matrix which has  the same non-zero eigenvalues as $F_{LN}$: 
\begin{align}
    F^*_{LN} &=Q(R-\bar{R})^\top(R-\bar{R}) = Q F^*_{mLN},
\end{align}
where
\begin{align}
F^*_{mLN} &:=(R-\bar{R})^\top (R-\bar{R}).
\end{align}
This $F^*_{mLN}$ corresponds to the mean subtraction in layer normalization. 
Its entries are given by 
\begin{align}
F^*_{mLN}(k,k')_{st} &:=(\nabla_\theta u_k^L(s) -\nabla_\theta \mu^L(s))^{\top}  (\nabla_\theta u_{k'}^L(t) -\nabla_\theta \mu^L(t)) \\
&= u_k^L(s)^{\top} \nabla_\theta u_{k'}^L(t) - \frac{\sum_a}{C}\nabla_\theta  u_k^L(s)^{\top} \nabla_\theta u_{a}^L(t) -\frac{\sum_a}{C} \nabla_\theta  u_{a}^L(s)^{\top} \nabla_\theta u_{k'}^L(t)  \nonumber \\ 
 &  + \frac{\sum_a}{C^2} \nabla_\theta u_a^L(s)^{\top} \nabla_\theta u_{a}^L(t) + \frac{\sum_{a\neq a'}}{C^2}  \nabla_\theta u_a^L(s)^{\top} \nabla_\theta u_{a'}^L(t).  
 \end{align}
In the large $M$ limit, we have 
\begin{align}
\nabla_\theta u_k^L(s)^{\top} \nabla_\theta u_{k'}^L(t)  
&=\sum_{l,ij} (\delta_{k,i}^l(s)\delta_{k',i}^l(t))(h_{j}^{l-1}(s)h_{j}^{l-1}(t)) \\
&= \sum_{l}  M_l   \tilde{q}^l_{st}\hat{q}^{l-1}_{st}\delta_{kk'}+O(M^{1-q^*}), 
\end{align}
after doing the same calculation as Eq. (\ref{eq12:0522}) and using the order parameters obtained in Section E.1. We have $q^*=1/2$ due to the gradient independence assumption. 
The reversed FIM becomes  
 \begin{equation}
    F^*_{mLN}(k,k')=\left(\delta_{kk'}-\frac{1}{C}\right)\alpha\frac{M}{T} K_{LN}   + \frac{1}{T}O(M^{1-q^*}),
\end{equation}
where we defined a matrix $K_{LN}$ by 
\begin{equation}
(K_{LN})_{st}=   \kappa'_1 \ \  (s=t), \ \   \kappa'_2 \ \ (s \neq t),
\end{equation}
with  
\begin{equation}
{\kappa'}_1 :=  \sum_{l=1}^L\ \frac{\alpha_{l-1}}{\alpha} \tilde{q}^l_t  \hat{q}^{l-1}_t, \ \ {\kappa'}_2 := \sum_{l=1}^L   \frac{\alpha_{l-1}}{\alpha} \tilde{q}^l_{st}\hat{q}^{l-1}_{st}.
\end{equation}
Note that the order parameters $(\hat{q}^l_t,\hat{q}_{st}^l,\tilde{q}^l_t,\tilde{q}_{st}^l)$ of layer normalization are computed by the recurrence relations (Eqs. (\ref{eq79:0517}) and (\ref{eq89:0517})).
Finally, the ($k,k'$)-th block of $F^*_{LN}$ is given by
\begin{align}
  F^*_{LN}(k,k') &= \sum_a Q(k,a)F^*_{mLN}(a,k') \\
  &= \alpha\frac{M}{T} \mathrm{diag} \left(\frac{1}{\sigma(t)^2} \right)  \left(\left(\delta_{kk'}-\frac{1}{C}\right)I_T-\frac{1}{C}\mathrm{diag} \left(\frac{\bar{u}_k(t)\bar{u}_{k'}(t)}{\sigma(t)^2} \right) \right)K_{LN}, \label{eq102:0517}
\end{align}
where $\mathrm{diag}(f(t))$ means a $T \times T$ diagonal matrix whose $t$-th diagonal entry is $f(t)$. 

\subsubsection{Eigenvalue statistics}
The mean is asymptotically given by
\begin{align}
m_\lambda &= \mathrm{Trace}(F^*_{LN})/P \\
&= \sum_{k} \mathrm{Trace}(F^*_{LN}(k,k))/P \\ 
&\sim \eta_1 \frac{(C-2) \kappa'_1}{M},
\end{align}
where $\eta_1 := \frac{1}{T}\sum_t \frac{1}{\sigma(t)^2}$. 

The largest eigenvalue is evaluated using the second moment of the eigenvalues. 
Since the second moment $s_\lambda=\sum_i\lambda_i^2/P$ is given by a trace of the squared matrix in general, we have  
\begin{align}
s_\lambda &= \mathrm{Trace}((F^*_{LN})^2)/P \\
&=  \sum_{k} \mathrm{Trace}(\sum_{a} F^*_{LN}(k,a)F^*_{LN}(a,k))/P. \label{eq109:0523} 
\end{align}
We obtain 
\begin{align}
 &((\sum_{a} F^*_{LN}(k,a)F^*_{LN}(a,k))_{tt} \\
 &=(\alpha\frac{M}{T})^2 \sum_{a,t'} \frac{1}{\sigma(t)^2} \left(\left(\delta_{ka}-\frac{1}{C}\right)-\frac{1}{C}\ \left(\frac{\bar{u}_k(t)\bar{u}_{a}(t)}{\sigma(t)^2} \right) \right)(K_{LN})_{tt'} \\ 
 &\ \ \ \ \cdot \frac{1}{\sigma(t')^2} \left(\left(\delta_{ak}-\frac{1}{C}\right)-\frac{1}{C} \left(\frac{\bar{u}_a(t')\bar{u}_{k}(t')}{\sigma(t')^2} \right) \right) (K_{LN})_{t't} \nonumber \\
&= \sum_{t'} (\alpha\frac{ M}{T})^2 \frac{1}{\sigma(t)^2\sigma(t')^2}  (C-3+g(t,t')^2)(({\kappa'_1}^2-{\kappa'_2}^2)\delta_{tt'}+{\kappa'_2}^2), \label{eq112:0523}
\end{align}
where we define $g(t,t'):= \frac{1}{C}\frac{\sum_a \bar{u}_a(t)\bar{u}_a(t') }{\sigma(t)\sigma(t')}.$ Substituting Eq. (\ref{eq112:0523}) into Eq. (\ref{eq109:0523}), we obtain
\begin{align}
s_\lambda &=  \alpha  \frac{ (\eta_3^2-\eta_1^2)T+(C-2)(\eta_1^2T-\eta_2)}{T}{\kappa'_2}^2 + \alpha (C-2)\eta_2\frac{{\kappa_1'}^2}{T},
\end{align}
where $\eta_2:=\frac{\sum_{t}}{T} \frac{1}{\sigma(t)^4} $ and $\eta_3:= \frac{1}{T^2}\sum_{t,t'}\frac{g(t,t') }{\sigma(t)\sigma(t')}.$
The lower bound is given by $\lambda_{max} \geq s_\lambda/m_\lambda$ and the upper bound by 
 $\lambda_{max} \leq \sqrt{Ps_\lambda}$.

{\bf Remark on $C=2$:} Because we have a special symmetry, i.e., $\bar{u}_1(t)=-\bar{u}_2(t)= (u_1^L(t)-u_2^L(t))/2$ in $C=2$, the gradient (Eq. (\ref{eq86:0517})) becomes zero. This is caused by the mean subtraction and variance normalization in the last layer. 
This makes the FIM a zero matrix. The case of $C>3$ is non-trivial and the FIM becomes non-zero, as we revealed.   
Similarly, the gradient (Eq. (\ref{eq55:0517})) in batch normalization becomes zero when $T=2$ due to the same symmetry \cite{yang2019bn}.  Such an exceptional case of batch normalization is not our interest because we focus on the sufficiently large $T$ in Eq. (\ref{eq19:190523}).

\end{document}